\documentclass[11pt]{article}
\usepackage[final]{acl}
\usepackage{times}
\usepackage{latexsym}
\usepackage[T1]{fontenc}
\usepackage[utf8]{inputenc}
\usepackage{microtype}
\usepackage{inconsolata}
\usepackage{graphicx}
\usepackage{booktabs}
\usepackage{amsmath}
\usepackage{tikz}
\usetikzlibrary{arrows.meta,positioning,fit,backgrounds,calc}
\definecolor{figctrl}{RGB}{0,114,178}
\definecolor{figctrlbg}{RGB}{229,240,247}
\definecolor{figevid}{RGB}{163,113,0}
\definecolor{figevidbg}{RGB}{251,240,217}
\definecolor{figfin}{RGB}{0,133,97}
\definecolor{figfinbg}{RGB}{223,243,237}
\definecolor{figmut}{RGB}{92,98,104}
\definecolor{figneut}{RGB}{243,244,246}
\usepackage{multirow}
\usepackage{array}
\usepackage{placeins}
\usepackage{balance}
\usepackage{needspace}

\hypersetup{
  pdftitle={Thakaa at AlexandriaX-2026: Who Gets the Final Say? Assigning Output Authority in Dialectal Arabic Machine Translation and Error Detection},
  pdfauthor={Meshal Alamr; Maryam Altuwaijri; Hassan Alqaeri; Abdullah Aldahlawi},
  pdfsubject={System description paper for the AlexandriaX-2026 shared task},
  pdfkeywords={dialectal Arabic machine translation; minimum Bayes risk; LoRA; in-context learning; translation error detection; AlexandriaX-2026; Thakaa}
}

\title{Thakaa at AlexandriaX-2026: Who Gets the Final Say? Assigning Output Authority in Dialectal Arabic Machine Translation and Error Detection}

\author{Meshal Alamr \quad Maryam Altuwaijri \quad Hassan Alqaeri \quad Abdullah Aldahlawi \\
  Thakaa, Advanced Company for Information Technology \\
  Riyadh, Saudi Arabia \\
  \texttt{\{m.alamr,m.altuwaijri,h.alqaeril,aldahlawi\}@thakaait.net}}

\begin{document}
\maketitle

\begin{abstract}
We report first-place systems for all four evaluated AlexandriaX-2026 settings: both English-to-dialect translation tracks, cross-dialect translation, and translation-error detection. In unconstrained translation, our minimum Bayes risk (MBR) selection restricts output authority: four local systems can supply the translation, while seven additional systems contribute only reference evidence. Candidate-constrained MBR beats the strongest complete candidate in all 13 dialects and gains 1.2554 macro spBLEU overall, reaching 33.4891. It also surpasses a retrospective oracle selecting the best whole candidate system per dialect. In the constrained track, country-routed LoRA adapters over a shared NileChat-3B-Base backbone incorporate the released Libyan and Sudanese supervision absent from the training and development splits, raising macro spBLEU from 29.5969 to 30.4233. Post-evaluation, a single adapter trained on all released Alexandria data reaches 30.4733 macro spBLEU; averaging model weights across three training seeds raises this to 30.9726 while retaining one model for inference. For cross-dialect translation, nonparallel target-dialect retrieval guides deliberation, raising Gemma-4-31B-IT without fine-tuning from 24.6898 to 28.4011 spBLEU. For translation-error detection, GPT-5.5 proposes semantic error regions, while MARBERTv2 controls their character boundaries and corroborates additional spans, raising Overall from 0.4691 to 0.4982.
\end{abstract}

\section{Introduction}

Dialectal Arabic systems must preserve meaning while realizing a specified target variety,
despite uneven supervision across dialects; translation-error detection additionally requires
exact character-level localization. AlexandriaX-2026 evaluates these challenges through
English-to-dialect conversational translation, cross-dialect translation, and character-level
detection and classification of MT errors \citep{elmekki-etal-2026-alexandriax}. We participated
in every task and both Subtask 1 tracks, placing first in all four evaluated settings
(Table~\ref{tab:results}).

\begin{table}[b]
\centering
\small
\setlength{\tabcolsep}{3.4pt}
\begin{tabular}{llrrr}
\toprule
Task & Setting & Primary & chrF++ & Rank \\
\midrule
ST1 & Constrained & 30.4233 & 45.4250 & \textbf{1/6} \\
ST1 & Unconstrained & 33.4891 & 48.1320 & \textbf{1/6} \\
ST2 & Cross-dialect & 28.4011 & 42.1509 & \textbf{1/4} \\
ST3 & Error detection & 0.4982 & n/a & \textbf{1/6} \\
\bottomrule
\end{tabular}
\caption{Official blind-test results. ST: subtask; rank: position among participating teams. Primary: spBLEU for ST1/ST2 (SacreBLEU, FLORES-200 tokenizer; \citealp{post-2018-call,nllbteam2022nllb}); Overall for ST3, averaging overlap F1 0.5426 and class F1 0.4538. ST1 averages 13 dialects; chrF++ \citep{popovic-2017-chrf} is secondary.}
\label{tab:results}
\end{table}

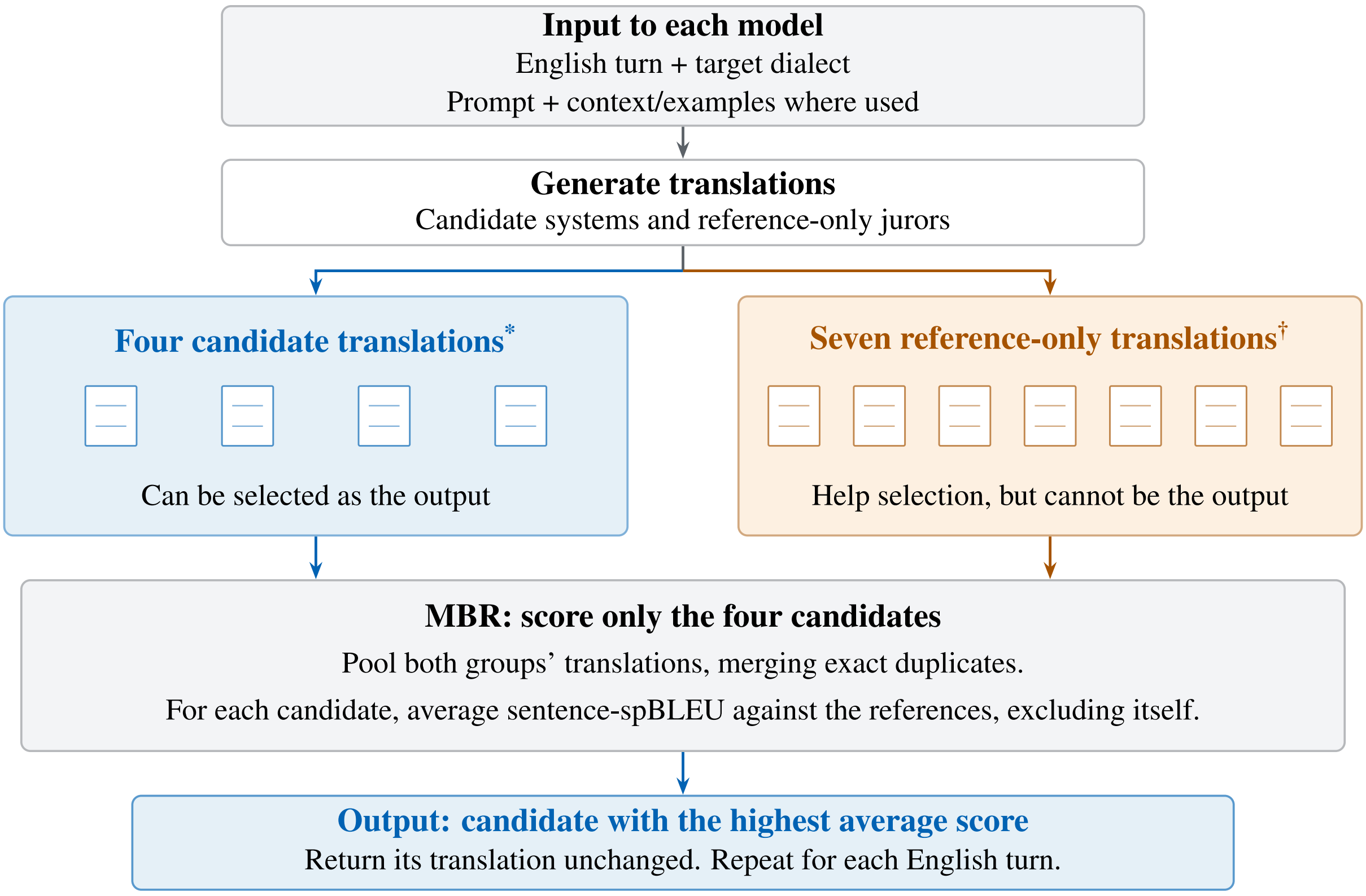
\begin{figure*}[!t]
\centering
\colorlet{figevid}{orange!65!black}
\definecolor{figevidbg}{RGB}{253,240,225}
\colorlet{figctrl}{blue!45!cyan!70!black}
\begin{tikzpicture}[x=1mm,y=1mm,
  txt/.style={font=\fontsize{10}{12}\selectfont,inner sep=0pt},
  title/.style={txt,font=\fontsize{11}{13}\selectfont\bfseries},
  box/.style={rounded corners=3pt,line width=.7pt,inner sep=0pt},
  translation/.style={rounded corners=.5pt,line width=.55pt,fill=white,minimum width=6mm,minimum height=7mm,inner sep=0pt},
  flow/.style={-{Stealth[length=2mm,width=1.5mm]},line width=.9pt}
]
\node[box,draw=figmut!45,fill=figneut,minimum width=108mm,minimum height=14mm] (input) at (79.5,0) {};
\node[title] at (79.5,4.5) {Input to each model};
\node[txt] at (79.5,0) {English turn + target dialect};
\node[txt] at (79.5,-4.5) {Prompt + context/examples where used};
\node[box,draw=figmut!45,fill=white,minimum width=108mm,minimum height=10mm] (generate) at (79.5,-16) {};
\node[title] at (79.5,-13.7) {Generate translations};
\node[txt] at (79.5,-18.3) {Candidate systems and reference-only jurors};
\draw[flow,figmut] (input.south) -- (generate.north);

\node[box,draw=figctrl!50,fill=figctrlbg,minimum width=73mm,minimum height=28mm] (candidates) at (36.5,-41) {};
\node[box,draw=figevid!50,fill=figevidbg,minimum width=73mm,minimum height=28mm] (jurors) at (122.5,-41) {};
\coordinate (branch) at (79.5,-24);
\draw[figmut,line width=.9pt] (generate.south) -- (branch);
\draw[flow,figctrl] (branch) -| (candidates.north);
\draw[flow,figevid] (branch) -| (jurors.north);
\node[title,text=figctrl] at ($(candidates.north)+(0,-5)$) {Four candidate translations\textsuperscript{*}};
\node[title,text=figevid] at ($(jurors.north)+(0,-5)$) {Seven reference-only translations\textsuperscript{\dag}};
\foreach \offset in {-24,-8,8,24} {
  \node[translation,draw=figctrl!70] (candidateDoc) at ($(candidates.center)+(\offset,0)$) {};
  \foreach \offsetY in {-1.2,1.2} \draw[figctrl!50,line width=.45pt] ($(candidateDoc.center)+(-1.8,\offsetY)$) -- ($(candidateDoc.center)+(1.8,\offsetY)$);
}
\foreach \offset in {-30,-20,-10,0,10,20,30} {
  \node[translation,draw=figevid!70] (jurorDoc) at ($(jurors.center)+(\offset,0)$) {};
  \foreach \offsetY in {-1.2,1.2} \draw[figevid!50,line width=.45pt] ($(jurorDoc.center)+(-1.8,\offsetY)$) -- ($(jurorDoc.center)+(1.8,\offsetY)$);
}
\node[txt] at ($(candidates.south)+(0,4.5)$) {Can be selected as the output};
\node[txt] at ($(jurors.south)+(0,4.5)$) {Help selection, but cannot be the output};

\coordinate (groupsBottom) at ($(candidates.south)!.5!(jurors.south)$);
\node[box,draw=figmut!45,fill=figneut,minimum width=155mm,minimum height=20mm,below=5mm of groupsBottom] (select) {};
\draw[flow,figctrl] (candidates.south) -- (select.north -| candidates.south);
\draw[flow,figevid] (jurors.south) -- (select.north -| jurors.south);
\node[title] at ($(select.center)+(0,5.5)$) {MBR: score only the four candidates};
\node[txt] at (select.center) {Pool both groups' translations, merging exact duplicates.};
\node[txt] at ($(select.center)+(0,-5.5)$) {For each candidate, average sentence-spBLEU against the references, excluding itself.};
\node[box,draw=figctrl!65,fill=figctrlbg,minimum width=129mm,minimum height=11mm,below=5mm of select] (output) {};
\draw[flow,figctrl] (select.south) -- (output.north);
\node[title,text=figctrl] at ($(output.center)+(0,2.3)$) {Output: candidate with the highest average score};
\node[txt] at ($(output.center)+(0,-2.3)$) {Return its translation unchanged. Repeat for each English turn.};
\end{tikzpicture}
\caption{Candidate-constrained MBR: additional translations help choose the answer, but only a candidate can supply it. For each turn, we select the candidate with the highest average sentence-spBLEU against the other unique translations (Eq.~\ref{eq:mbr}). \textsuperscript{*}Candidate models and training recipes: Table~\ref{tab:candidates}. \textsuperscript{\dag}Reference-only jury models and prompts: Appendix~\ref{app:repro}. The submitted system reaches 33.4891 macro spBLEU, 1.2554 above the best single candidate system and 0.8145 above choosing the best whole system separately for each dialect (32.6746). This retrospective comparator uses blind-test scores to choose one system for all turns in each dialect; MBR selects separately for each turn without hidden references (Section~\ref{sec:st1unconstrained}).}
\label{fig:st1}
\end{figure*}

The resulting systems combine fine-tuned generators, retrieved examples, selectors, and span
controllers in different ways. We use \emph{output authority} to distinguish components allowed
to emit or finalize the scored prediction from those that only condition, propose, or score it.
In unconstrained translation, extra systems help select an answer without being allowed to supply it (Figure~\ref{fig:st1}).

Controlled comparisons yield several findings. In unconstrained translation, the reference-only jury improves candidate-constrained MBR, whereas making the same systems selectable forfeits part of that gain. For NileChat-3B-Base, beam search produces the largest measured inference-time improvement, and authentic Libyan and Sudanese examples outperform an equal-size duplicate-row control. A separate subdialect-conditioning comparison shows that derived labels help chiefly where country-level labels collapse several annotated varieties. In cross-dialect translation, nonparallel target-dialect retrieval improves spBLEU, with a controlled gain from 16 to 32 examples. \par\begin{samepage}
In translation-error detection, missed spans and imprecise boundaries dominate the remaining errors, rather than category selection once a match exists. \mbox{Consistent} with this diagnosis, the blind-test gains come from encoder-guided boundary widening and agreement-gated span additions while retaining GPT-5.5 as the semantic proposer.
\par\end{samepage}

\section{Subtask 1: English-to-Dialect Translation}

Subtask 1 translates each English conversational turn into one of 13 country-specified Arabic
varieties. Inputs include conversation metadata and preceding dialogue context; the 14,459-turn
blind test is ranked by macro spBLEU across countries
\citep{elmekki-etal-2026-alexandriax,el-mekki-etal-2026-alexandria}. We address
complementary candidate selection in the unconstrained track and constrained coverage for Libyan
(LY) and Sudanese (SD) Arabic, which are absent from the official training and development splits.

\begin{table*}[t]
\centering
\small
\setlength{\tabcolsep}{5pt}
\begin{tabular}{l>{\raggedright\arraybackslash}p{0.673\textwidth}r}
\toprule
Candidate system & Adaptation, data, and decoding & spBLEU \\
\midrule
Gemma-4-12B-IT CPO & rank-16 QLoRA SFT: train+dev+public test + AraBench (2,500 LY + 2,000 SD examples); CPO: 13,049 chrF++-ranked gold-vs-best pairs; terminal-punctuation normalization & 32.0318 \\
\addlinespace[2pt]
12B CPO + ICL & same trained 12B weights + $k{=}8$ Alexandria training-split examples where available (none for LY/SD), retrieved by identifier group (BM25 fallback) & 32.2235 \\
\addlinespace[2pt]
Gemma-4-E4B-IT SFT & rank-32 LoRA SFT: train + public test + AraBench LY/SD (no dev); beam-8 & 32.2336 \\
\addlinespace[2pt]
NileChat-3B-Base soup & $3\times$ rank-32 LoRA (train+dev), uniform weight average; beam-8, MR/MA training-split demonstrations & 29.5969 \\
\bottomrule
\end{tabular}
\caption{The four systems producing the selectable translations in Figure~\ref{fig:st1}, with their adaptation recipes and official standalone blind-test spBLEU \citep{gemmateam2026gemma4technicalreport,pmlr-v235-xu24t,sajjad-etal-2020-arabench,el-mekki-etal-2025-nilechat}. SFT: supervised fine-tuning; CPO: contrastive preference optimization; ICL: in-context learning. All use metadata-derived subdialect labels. The NileChat row is the original train+development soup, not the routed constrained submission or the post-evaluation full-data soup.}
\label{tab:candidates}
\end{table*}

\subsection{Shared conditioning: metadata-derived subdialect labels}
\label{sec:subdialect}

The Alexandria dataset study reports stable within-country subdialect difficulty across models but
mixed gains from general metadata \citep{el-mekki-etal-2026-alexandria}. We isolate the target
label itself: training and development provide fine-grained annotations, whereas blind-test inputs
generally identify only the country. In training, identifier prefixes map deterministically to the
subdialect in all 129 observed country-prefix pairs. The mapping labels 98\% of blind-test
conversations in the eleven represented countries, uses a country-level fallback elsewhere, and
changes the country label in 24\% of prompts. All submitted Subtask 1 systems use this benchmark-specific label.

Between two near-matched official submissions of our strongest 12B candidate (same model and greedy decoding; different batch sizes), switching the prompt from country to subdialect raises macro spBLEU from 30.61 to 31.79 (+1.18): the five multi-subdialect countries supply +1.14 (OM +7.47 to SA +1.01) and the other eight only +0.04, suggesting that batch-size drift is small. In an Oman-only development ablation, the subdialect-trained adapter improves only when the matching fine-grained label is also supplied at inference (Figure~\ref{fig:conditioning}(a)). These comparisons support preserving target-label granularity when labels are available. The identifier-prefix mapping is benchmark-specific; these experiments evaluate supplied or metadata-derived labels, not subdialect identification from text.

\Needspace{5\baselineskip}
\subsection{Unconstrained system: candidate-constrained MBR}
\label{sec:st1unconstrained}

The candidates follow separate improvement paths (Table~\ref{tab:candidates};
Appendix~\ref{app:repro}): contrastive
preference optimization (CPO) on a fine-tuned Gemma-4-12B-IT; identifier-matched demonstrations
for its in-context-learning (ICL) variant \citep{agrawal-etal-2023-context}; fold-in supervised fine-tuning (SFT), which adds released public-test rows, and beam search for the compact E4B (effective 4B); and the constrained-track
NileChat soup. Their outputs are complementary: post-evaluation demotion of any candidate costs
0.1332 to 0.4175 macro spBLEU, and the soup wins 12.6\% of turns despite its lowest standalone
score, so all four supply useful output hypotheses (Appendix~\ref{app:st1jury}).

For each turn, the candidates form the selectable set $\mathcal{C}(x)$; a fine-tuned Gemma-4-E2B-IT system
(effective 2B) and six prompted GPT/Gemini systems form a reference-only jury $\mathcal{J}(x)$.
Exact duplicates across all eleven outputs collapse into single hypotheses, giving one pool
$\mathcal{H}(x)$; with $\mathcal{H}_y=\mathcal{H}(x)\setminus\{y\}$, we select
\begin{equation}
\hat{y}=\underset{y\in\mathcal{C}(x)}{\arg\max}\;
\frac{1}{|\mathcal{H}_y|}\sum_{r\in\mathcal{H}_y}
\operatorname{spBLEU}(y,r).
\label{eq:mbr}
\end{equation}
Each distinct juror hypothesis can affect agreement, while only a local candidate can be returned. The rule follows MBR decoding and system combination \citep{kumar-byrne-2004-minimum,eikema-aziz-2022-sampling,freitag-etal-2022-high,gonzalez-rubio-casacuberta-2013-improving,farinhas-etal-2023-empirical}. Unlike recombination or preference optimization \citep{jon-bojar-2023-breeding,wang-etal-2026-m2po}, it selects among fixed outputs. Building on multi-prompt and dialectal-Arabic MBR \citep{heineman-etal-2024-improving,alali-issam-2026-maastricht}, we evaluate restricting outputs to the four local candidates.

\paragraph{Selection versus additional reference evidence.}
With the four candidate outputs fixed, a post-evaluation control using only those candidates as mutual references scores 33.0925, exceeding the strongest standalone candidate, E4B at 32.2336, by 0.8589. Adding the seven reference-only jurors raises this to the submitted 33.4891, a further 0.3966. Of that jury increment, E2B adds 0.1025 before the API jurors are introduced; the six API jurors then add 0.2940 and improve 10 of 13 countries. Thus, most of the 1.2554-point gain is already obtained by selection within the local pool, with additional reference evidence improving that selection. This is a conditional score decomposition for fixed candidates and utility, not an independent attribution to each model: the jury-family factorial shows a positive interaction (Appendix~\ref{app:st1jury}). Appendix~\ref{app:comparisonrecord} consolidates the main comparisons across all four settings, identifying the reference configurations, score changes, and evaluation phases.

The selector also scores 0.8145 above a retrospective dialect-level system oracle that selects one complete candidate per dialect using blind-test scores (32.6746); its margin is positive in all 13 dialects, demonstrating within-dialect \mbox{complementarity} (Table~\ref{tab:st1bydialect}). In a post-evaluation control, granting the jurors output eligibility over the same hypothesis pool and utility scores 33.2998, forfeiting 0.1893 of the gain (Figure~\ref{fig:authority}).

In a separate post-evaluation experiment, the MBR system with the full-data soup (Section~\ref{sec:constrained}) replacing the original NileChat candidate reaches 33.5246 macro spBLEU, slightly above the submitted system's official 33.4891 (Appendix~\ref{app:st1jury}).

\subsection{Constrained system: shared-backbone LoRA routing}
\label{sec:constrained}

The constrained track permits only Alexandria data and caps the system at 5B raw parameters.
Track policy admits all released data; the public-test split is the only released LY/SD
supervision. Using the released country field, we route LoRA adapters \citep{hu-etal-2021-lora}
over one shared 3.086B \texttt{NileChat-3B-Base}: the default is the uniform three-seed soup of
Table~\ref{tab:candidates} \citep{pmlr-v162-wortsman22a}, while LY/SD use one same-recipe adapter
trained on train plus public test.

Both routes decode with beam-8 in bf16, condition on the derived subdialect label
(Section~\ref{sec:subdialect}), and reuse preceding Arabic outputs as self-context. Mauritanian
(MR) and Moroccan (MA) inputs also receive three same-dialect training-split exemplars in the
training template (+0.85/+0.81 development spBLEU; EG, TN, and YE controls flat or worse). The
backbone and four deltas total approximately 3.33B parameters. Routing changes only the 2,224 LY/SD
turns, leaving the other eleven countries byte-identical, so the gain from 29.5969 to 30.4233
comes entirely from LY (21.85 to 26.72) and SD (22.29 to 28.17)
(Table~\ref{tab:st1percountry}; Figure~\ref{fig:conditioning}(b)).

A matched post-evaluation ablation isolates the value of authentic LY/SD supervision (Appendix~\ref{app:st1constrained}). Its baseline is a single seed-42 adapter trained on training and development data, not the three-seed soup used in the submission. Adding public-test LY/SD raises macro spBLEU from 29.2323 to 30.1179, while an equal-size duplicate-row control falls to 29.0971. Training on all released Alexandria data reaches 30.4733, versus 28.8483 for full-budget duplication, supporting the value of new supervision rather than additional optimizer updates alone.

\paragraph{Weight averaging after full-data training.}
A further post-evaluation comparison repeats the full-data recipe with seeds 42, 123, and 7. Individual models score 30.4426 to 30.4733; averaging their merged weights yields 30.9726, a gain of 0.4993 over the best seed, and improves spBLEU over the strongest individual seed in every dialect (Table~\ref{tab:st1seeds}). This soup is decoded as one 3.086B-parameter model, without routing or an inference-time ensemble. It is distinct from the submitted train+development soup. Together, full-data training and weight averaging give a stronger single-model recipe than the routed submission, although the different training configurations do not isolate the effect of removing routing. Further beam-search and conditioning controls appear in Appendix~\ref{app:st1constrained}.

\section{Subtask 2: Cross-Dialect Translation}

Subtask 2 comprises 3,574 retail-banking translations across all 30 directed pairs among six Arabic
dialects \citep{elmekki-etal-2026-alexandriax,malaysha-etal-2024-arafinnlp}. No task-specific
training or development references
are released, leaving target terminology, morphology, and spelling underdetermined even when the
banking intent is preserved.

\subsection{System: target-dialect retrieval and guided generation}

For a query $q$ from dialect $A$ to $B$, the retrieval bank contains released blind inputs written
in $B$, treated as unpaired target-dialect usage examples with no cross-dialect alignment used in
retrieval. We embed unique sources with
multilingual E5-Large \citep{wang2024multilinguale5textembeddings}. To suppress likely
cross-dialect counterparts, we discard candidate $c$ when $q$ is among its 32 nearest dialect-$A$
sources, then rank survivors by a bitext-mining relative margin
(\citealp{artetxe-schwenk-2019-margin}; Equation~\ref{eq:st2margin}). The top 32 enter the prompt as
usage evidence; none is identified or presented as a translation of $q$
\citep{tan-etal-2024-narrowing}.

A fixed prompt asks a frozen Gemma-4-31B-IT, decoding greedily, to weigh source-faithful,
target-idiomatic, and retrieval-aligned renderings into one synthesis, a single-call latent variant
of reasoning prompts for MT \citep{nguyen-xu-2025-reasoning}. The pipeline never queries hidden reference data or intent labels. An exact-copy guard regenerates whenever the normalized output copies an active
example, falling back to the clean no-retrieval output on persistent copies
(Appendix~\ref{app:st2retrieval}).

\subsection{Results and integrity analysis}

The retrieved messages expose target-dialect vocabulary, morphology, spelling, and style.
Consistent with the model exploiting this evidence, official spBLEU rises from 24.6898 without retrieval to 27.9496 at $k{=}32$; the 16-to-32 comparison holds copy handling fixed (Appendix~\ref{app:st2ladder}). Guided generation brings the final system to 28.4011 (+0.45; Table~\ref{tab:st2ladder}).

The method is transductive over the released blind sources. A retrospective audit against public
ArBanking77 counterparts \citep{jarrar-etal-2023-arbanking77} found at least two recoverable query counterparts among 3,574 final prompts, including one case of wording reuse. We exclude
overlapping ArBanking77 rows from independent evaluation claims; Appendix~\ref{app:st2audit} documents
the audit and exploratory analyses.

\section{Subtask 3: Error Span Detection and Classification}

Subtask 3 requires translation-error localization and six-way classification under Linguistically Motivated Multidimensional Quality Metrics (LQM)
\citep{magdy-etal-2026-lqm} over 1,125 training, 138 development, and 145 blind-test examples in
five directions \citep{elmekki-etal-2026-alexandriax}. Overall averages character-overlap F1 and pooled error-class F1, so systems must
both localize errors and label matched spans; the task sits between LLM error analysis
\citep{kocmi-federmann-2023-gemba} and span-supervised evaluation \citep{guerreiro-etal-2024-xcomet}.

\subsection{System: semantic proposal and span control}

Given the English source and its Arabic translation, GPT-5.5 proposes semantic error regions by
reproducing the translation with inline category tags, which we map deterministically to
character offsets. A 162M-parameter MARBERTv2 encoder \citep{abdul-mageed-etal-2021-arbert},
trained as a joint 13-label BIO tagger on the 1,125 training rows with the 138 development rows
used for epoch selection, widens any overlapping proposal to the encoder boundary, since the
overlap half of the score counts characters while an LLM tends to mark only the word it judges
wrong. Gemini 3.6 Flash may also widen spans; a disjoint Gemini proposal is admitted only when
MARBERTv2 independently predicts an overlapping error. Rows left without a prediction use a fixed
fallback chain, and adjacent same-category fragments separated by at most one character are
merged. The translation itself is never edited; Appendix~\ref{app:repro} records implementation
and prompt details.

\subsection{Results and error analysis}

The official blind-test ladder records the contribution of each retained stage: raw GPT-5.5 scores 0.4691 Overall, encoder-guided widening reaches 0.4786, extended widening with agreement-gated additions and empty-row rescue reaches 0.4967, and fragment repair produces the final 0.4982 (Table~\ref{tab:st3all}). The best fully local alternative combines Gemma-4-31B-IT and MARBERTv2 at character level and reaches 0.4446. In pre-test cross-validation of a direction-routed system, 51.7\% of predicted spans match no gold span and 33.7\% of gold spans are missed, whereas category accuracy reaches 75.35\% once a span is matched (Figure~\ref{fig:st3panels}). In these predictions, finding and localizing errors is therefore a larger bottleneck than classifying an error after it has been found.

\section{Conclusion}

We developed first-place systems for all four AlexandriaX-2026 settings, with matched controls identifying where their gains came from. In the constrained track, authentic Libyan and Sudanese supervision outperforms matched duplicate-row controls. Post-evaluation, averaging three full-data models improves spBLEU over every individual seed in all 13 dialects, yielding a stronger single-model recipe without routing. Source-only target-dialect retrieval improves generation without fine-tuning, while error detection improves when semantic proposal and span control are separated.

Across these systems, component value does not simply follow parameter count. Among the submitted candidates, compact E4B is the strongest standalone generator and largest conditional contributor, yet turn-level MBR surpasses every candidate in all 13 dialects. A 162M-parameter encoder likewise improves the boundaries of frontier-model proposals. The authority control separates evidence from generation: reference-only jurors strengthen consensus, but making them selectable forfeits 0.1893 spBLEU. Together, these results support assigning each component its empirically strongest role.

\section*{Limitations}

The Alexandria dataset study underlying Subtask 1 finds that current models preserve meaning more
reliably than they produce authentic local dialect \citep{el-mekki-etal-2026-alexandria}. Our
official results rely on single-reference spBLEU and chrF++, which cannot determine whether a gain
reflects greater dialect authenticity or closer agreement with one translator's wording. We did
not conduct a separate human evaluation and therefore interpret these scores as benchmark
performance rather than direct evidence of naturalness.

Two design choices are specific to the released benchmark. The subdialect mapping depends on
patterns in the conversation identifiers, and the constrained fold-in adapter uses the only
released LY/SD supervision, leaving no held-out Alexandria split for these dialects. Subtask 2 is
also transductive. Because no parallel task data were released, retrieval uses blind-source
sentences written in the target dialect to expose vocabulary, morphology, spelling, and style
without identifying them as translations. The same pool can nevertheless contain semantically
corresponding sentences across dialects. Reciprocal-neighbor filtering removed likely alignments,
but a retrospective audit found at least two recoverable counterparts among 3,574 final prompts, with coverage limited to public data. The Subtask
2 result therefore characterizes retrieval-primed inference over this released pool rather than
conventional inductive generalization.

For the blind test, unreleased references preclude confidence intervals over official deltas. We
therefore report the smallest Subtask 3 increments as observed pipeline stages rather than stable
effect estimates. Controls designed after leaderboard feedback are marked post-evaluation and
treated as diagnostic; Appendix~\ref{app:st3validation} reports the Subtask 3 development-to-test analysis.

Finally, the unconstrained Subtask 1 jury and the Subtask 3 proposal pipeline use proprietary API
models. Their outputs were cached, but exact regeneration depends on continued access to the
reported endpoints.

\section*{Acknowledgments}

We thank the AlexandriaX-2026 organizers for providing the task resources and evaluation
infrastructure. We are especially grateful for their professionalism and exemplary responsiveness
throughout the shared task.

\bibliography{refs}

@inproceedings{elmekki-etal-2026-alexandriax,
    title = "{AlexandriaX 2026: The First Shared Task on Dialectal Arabic Machine Translation}",
    author = "El Mekki, Abdellah  and
      Elmadany, AbdelRahim A.  and
      Magdy, Samar M.  and
      Ezzini, Saad  and
      El-Haj, Mo  and
      Jarrar, Mustafa  and
      Alyafeai, Zaid  and
      Ghanem, Bernard  and
      Abdul-Mageed, Muhammad",
    booktitle = "Proceedings of The Fourth Arabic Natural Language Processing Conference: Shared Tasks",
    year = "2026",
    address = "Budapest, Hungary",
    publisher = "Association for Computational Linguistics",
}

@inproceedings{el-mekki-etal-2026-alexandria,
  title = {Alexandria: A Multi-Domain Dialectal {A}rabic Machine Translation Dataset for Culturally Inclusive and Linguistically Diverse {LLM}s},
  author = {El Mekki, Abdellah and Magdy, Samar M. and Atou, Houdaifa and AbuHweidi, Ruwa and Qawasmeh, Baraah and Nacar, Omer and Al-hibiri, Thikra and Saadie, Razan and Alsayadi, Hamzah A. and Hammouda, Nadia Ghezaiel and Alkhazimi, Alshima Mohammed and Hamod, Aya and Al-Ghafri, Al-Yas Yaqoob and El-Sayed, Wesam and al Sharji, Asila Ismail and Ballout, Mohamad and Belfathi, Anas and Ghaddar, Karim and Sibaee, Serry and Aoun, Alaa and Aseri, Aeej Mohammed and Abureesh, Lina and Bashiti, Ahlam and Yousef, Majdal and Hafiz, Abdulaziz and Mohamed, Yehdih and Hamedtou, Emira and Emehah, Brakehe and Alhamouri, Rahaf and Nafea, Youssef and El Aatar, Aya and Al-Dhabyani, Walid and Hamed, Emhemed S. and Shatnawi, Sara and Alwajih, Fakhraddin and Elkhidir, Khalid and Alasmari, Ashwag and Gerrio, Abdurrahman and Alshahri, Omar Said and Elmadany, AbdelRahim A. and Berrada, Ismail and Al-kathiri, Amir Azad Adli and Zaraket, Fadi and Jarrar, Mustafa and El Hadj, Yahya Mohamed and Alhuzali, Hassan and Abdul-Mageed, Muhammad},
  booktitle = {Proceedings of the 64th Annual Meeting of the Association for Computational Linguistics (Volume 1: Long Papers)},
  year = {2026},
  address = {San Diego, California, United States},
  publisher = {Association for Computational Linguistics},
  url = {https://aclanthology.org/2026.acl-long.1503/},
  doi = {10.18653/v1/2026.acl-long.1503},
  pages = {32567--32592}
}

@inproceedings{el-mekki-etal-2025-nilechat,
  title     = {{N}ile{C}hat: Towards Linguistically Diverse and Culturally Aware {LLM}s for Local Communities},
  author    = {El Mekki, Abdellah and Atou, Houdaifa and Nacar, Omer and Shehata, Shady and Abdul-Mageed, Muhammad},
  booktitle = {Proceedings of the 2025 Conference on Empirical Methods in Natural Language Processing},
  year      = {2025},
  address   = {Suzhou, China},
  publisher = {Association for Computational Linguistics},
  pages     = {10967--10991},
  doi       = {10.18653/v1/2025.emnlp-main.556},
  url       = {https://aclanthology.org/2025.emnlp-main.556/}
}

@inproceedings{magdy-etal-2026-lqm,
    title = "{LQM}: Linguistically Motivated Multidimensional Quality Metrics for Machine Translation",
    author = "Magdy, Samar M.  and
      Alwajih, Fakhraddin  and
      El Mekki, Abdellah  and
      El Sayed, Wesam  and
      Abdul-Mageed, Muhammad",
    booktitle = "Findings of the {A}ssociation for {C}omputational {L}inguistics: {ACL} 2026",
    month = jul,
    year = "2026",
    address = "San Diego, California, United States",
    publisher = "Association for Computational Linguistics",
    url = "https://aclanthology.org/2026.findings-acl.2012/",
    doi = "10.18653/v1/2026.findings-acl.2012",
    pages = "40470--40493"
}

@inproceedings{malaysha-etal-2024-arafinnlp,
    title = "{A}ra{F}in{NLP} 2024: The First {A}rabic Financial {NLP} Shared Task",
    author = "Malaysha, Sanad  and
      El-Haj, Mo  and
      Ezzini, Saad  and
      Khalilia, Mohammed  and
      Jarrar, Mustafa  and
      Almujaiwel, Sultan  and
      Berrada, Ismail  and
      Bouamor, Houda",
    booktitle = "Proceedings of the Second Arabic Natural Language Processing Conference",
    month = aug,
    year = "2024",
    address = "Bangkok, Thailand",
    publisher = "Association for Computational Linguistics",
    url = "https://aclanthology.org/2024.arabicnlp-1.34/",
    doi = "10.18653/v1/2024.arabicnlp-1.34",
    pages = "393--402"
}

@inproceedings{jarrar-etal-2023-arbanking77,
  title = {{A}r{B}anking77: Intent Detection Neural Model and a New Dataset in Modern and Dialectical {A}rabic},
  author = {Jarrar, Mustafa and Birim, Ahmet and Khalilia, Mohammed and Erden, Mustafa and Ghanem, Sana},
  booktitle = {Proceedings of ArabicNLP 2023},
  year = {2023},
  address = {Singapore (Hybrid)},
  publisher = {Association for Computational Linguistics},
  url = {https://aclanthology.org/2023.arabicnlp-1.22/},
  doi = {10.18653/v1/2023.arabicnlp-1.22},
  pages = {276--287}
}

@InProceedings{pmlr-v235-xu24t,
  title = {Contrastive Preference Optimization: Pushing the Boundaries of {LLM} Performance in Machine Translation},
  author = {Xu, Haoran and Sharaf, Amr and Chen, Yunmo and Tan, Weiting and Shen, Lingfeng and Van Durme, Benjamin and Murray, Kenton and Kim, Young Jin},
  booktitle = {Proceedings of the 41st International Conference on Machine Learning},
  pages = {55204--55224},
  year = {2024},
  volume = {235},
  series = {Proceedings of Machine Learning Research},
  publisher = {PMLR},
  url = {https://proceedings.mlr.press/v235/xu24t.html}
}

@inproceedings{eikema-aziz-2022-sampling,
    title = "Sampling-Based Approximations to Minimum {B}ayes Risk Decoding for Neural Machine Translation",
    author = "Eikema, Bryan  and
      Aziz, Wilker",
    booktitle = "Proceedings of the 2022 Conference on Empirical Methods in Natural Language Processing",
    month = dec,
    year = "2022",
    address = "Abu Dhabi, United Arab Emirates",
    publisher = "Association for Computational Linguistics",
    url = "https://aclanthology.org/2022.emnlp-main.754/",
    doi = "10.18653/v1/2022.emnlp-main.754",
    pages = "10978--10993"
}

@inproceedings{gonzalez-rubio-casacuberta-2013-improving,
    title = "Improving the minimum {B}ayes' risk combination of machine translation systems",
    author = "Gonz{\'a}lez-Rubio, Jes{\'u}s and Casacuberta, Francisco",
    booktitle = "Proceedings of the 10th International Workshop on Spoken Language Translation: Papers",
    year = "2013",
    address = "Heidelberg, Germany",
    url = "https://aclanthology.org/2013.iwslt-papers.4/"
}

@inproceedings{jon-bojar-2023-breeding,
    title = "Breeding Machine Translations: Evolutionary approach to survive and thrive in the world of automated evaluation",
    author = "Jon, Josef and Bojar, Ond{\v{r}}ej",
    booktitle = "Proceedings of the 61st Annual Meeting of the Association for Computational Linguistics (Volume 1: Long Papers)",
    month = jul,
    year = "2023",
    address = "Toronto, Canada",
    publisher = "Association for Computational Linguistics",
    url = "https://aclanthology.org/2023.acl-long.122/",
    doi = "10.18653/v1/2023.acl-long.122",
    pages = "2191--2212"
}

@inproceedings{wang-etal-2026-m2po,
    title = "{M}$^2${PO}: Multi-Perspective Multi-Pair Preference Optimization for Machine Translation",
    author = "Wang, Hao and Xu, Linlong and Liu, Heng and Liu, Yangyang and Zhao, Xiaohu and Zeng, Bo and Shao, Liangying and Dong, Yichen and Wu, Xinwei and Zhou, Jiang and Dong, Tianyu and Zeng, Xiangxiang and Wang, Longyue and Luo, Weihua",
    booktitle = "Proceedings of the 64th Annual Meeting of the Association for Computational Linguistics (Volume 1: Long Papers)",
    month = jul,
    year = "2026",
    address = "San Diego, California, United States",
    publisher = "Association for Computational Linguistics",
    url = "https://aclanthology.org/2026.acl-long.469/",
    doi = "10.18653/v1/2026.acl-long.469",
    pages = "10315--10336"
}

@misc{wang2024multilinguale5textembeddings,
  title = {Multilingual {E5} Text Embeddings: A Technical Report},
  author = {Wang, Liang and Yang, Nan and Huang, Xiaolong and Yang, Linjun and Majumder, Rangan and Wei, Furu},
  year = {2024},
  eprint = {2402.05672},
  archivePrefix = {arXiv},
  url = {https://arxiv.org/abs/2402.05672}
}

@inproceedings{artetxe-schwenk-2019-margin,
  title = {Margin-based Parallel Corpus Mining with Multilingual Sentence Embeddings},
  author = {Artetxe, Mikel and Schwenk, Holger},
  booktitle = {Proceedings of the 57th Annual Meeting of the Association for Computational Linguistics},
  year = {2019},
  address = {Florence, Italy},
  publisher = {Association for Computational Linguistics},
  url = {https://aclanthology.org/P19-1309/},
  doi = {10.18653/v1/P19-1309},
  pages = {3197--3203}
}

@misc{gemmateam2026gemma4technicalreport,
  title = {Gemma 4 Technical Report},
  author = {{Gemma Team}},
  year = {2026},
  eprint = {2607.02770},
  archivePrefix = {arXiv},
  url = {https://arxiv.org/abs/2607.02770}
}

@inproceedings{kocmi-federmann-2023-gemba,
  title     = {{GEMBA-MQM}: Detecting Translation Quality Error Spans with {GPT-4}},
  author    = {Kocmi, Tom and Federmann, Christian},
  booktitle = {Proceedings of the Eighth Conference on Machine Translation (WMT)},
  month     = dec,
  year      = {2023},
  address   = {Singapore},
  pages     = {768--775},
  publisher = {Association for Computational Linguistics},
  url       = {https://aclanthology.org/2023.wmt-1.64/},
  doi       = {10.18653/v1/2023.wmt-1.64}
}

@article{guerreiro-etal-2024-xcomet,
  title   = {x{COMET}: Transparent Machine Translation Evaluation through Fine-grained Error Detection},
  author  = {Guerreiro, Nuno M. and Rei, Ricardo and van Stigt, Daan and Coheur, Luisa and Colombo, Pierre and Martins, Andr{\'e} F. T.},
  journal = {Transactions of the Association for Computational Linguistics},
  volume  = {12},
  pages   = {979--995},
  year    = {2024},
  address = {Cambridge, MA},
  publisher = {MIT Press},
  url     = {https://aclanthology.org/2024.tacl-1.54/},
  doi     = {10.1162/tacl_a_00683}
}

@inproceedings{abdul-mageed-etal-2021-arbert,
  title     = {{ARBERT} \& {MARBERT}: Deep Bidirectional Transformers for {A}rabic},
  author    = {Abdul-Mageed, Muhammad and Elmadany, AbdelRahim and Nagoudi, El Moatez Billah},
  booktitle = {Proceedings of the 59th Annual Meeting of the Association for Computational Linguistics and the 11th International Joint Conference on Natural Language Processing (Volume 1: Long Papers)},
  month     = aug,
  year      = {2021},
  address   = {Online},
  pages     = {7088--7105},
  publisher = {Association for Computational Linguistics},
  url       = {https://aclanthology.org/2021.acl-long.551/},
  doi       = {10.18653/v1/2021.acl-long.551}
}

@inproceedings{sajjad-etal-2020-arabench,
    title = "{A}ra{B}ench: Benchmarking Dialectal {A}rabic-{E}nglish Machine Translation",
    author = "Sajjad, Hassan  and
      Abdelali, Ahmed  and
      Durrani, Nadir  and
      Dalvi, Fahim",
    booktitle = "Proceedings of the 28th International Conference on Computational Linguistics",
    month = dec,
    year = "2020",
    address = "Barcelona, Spain (Online)",
    publisher = "International Committee on Computational Linguistics",
    url = "https://aclanthology.org/2020.coling-main.447/",
    doi = "10.18653/v1/2020.coling-main.447",
    pages = "5094--5107"
}

@inproceedings{kumar-byrne-2004-minimum,
    title = "Minimum {B}ayes-Risk Decoding for Statistical Machine Translation",
    author = "Kumar, Shankar  and
      Byrne, William",
    booktitle = "Proceedings of the Human Language Technology Conference of the North {A}merican Chapter of the Association for Computational Linguistics: {HLT}-{NAACL} 2004",
    year = "2004",
    address = "Boston, Massachusetts, USA",
    publisher = "Association for Computational Linguistics",
    url = "https://aclanthology.org/N04-1022/",
    pages = "169--176"
}

@article{freitag-etal-2022-high,
    title = "High Quality Rather than High Model Probability: Minimum {B}ayes Risk Decoding with Neural Metrics",
    author = "Freitag, Markus  and
      Grangier, David  and
      Tan, Qijun  and
      Liang, Bowen",
    journal = "Transactions of the Association for Computational Linguistics",
    volume = "10",
    year = "2022",
    address = "Cambridge, MA",
    publisher = "MIT Press",
    url = "https://aclanthology.org/2022.tacl-1.47/",
    doi = "10.1162/tacl_a_00491",
    pages = "811--825"
}

@inproceedings{farinhas-etal-2023-empirical,
    title = "An Empirical Study of Translation Hypothesis Ensembling with Large Language Models",
    author = "Farinhas, Ant{\'o}nio  and
      de Souza, Jos{\'e} G. C.  and
      Martins, Andr{\'e} F. T.",
    booktitle = "Proceedings of the 2023 Conference on Empirical Methods in Natural Language Processing",
    month = dec,
    year = "2023",
    address = "Singapore",
    publisher = "Association for Computational Linguistics",
    url = "https://aclanthology.org/2023.emnlp-main.733/",
    doi = "10.18653/v1/2023.emnlp-main.733",
    pages = "11956--11970"
}

@InProceedings{pmlr-v162-wortsman22a,
  title = {Model soups: averaging weights of multiple fine-tuned models improves accuracy without increasing inference time},
  author = {Wortsman, Mitchell and Ilharco, Gabriel and Gadre, Samir Ya and Roelofs, Rebecca and Gontijo-Lopes, Raphael and Morcos, Ari S and Namkoong, Hongseok and Farhadi, Ali and Carmon, Yair and Kornblith, Simon and Schmidt, Ludwig},
  booktitle = {Proceedings of the 39th International Conference on Machine Learning},
  pages = {23965--23998},
  year = {2022},
  volume = {162},
  series = {Proceedings of Machine Learning Research},
  publisher = {PMLR},
  url = {https://proceedings.mlr.press/v162/wortsman22a.html}
}

@inproceedings{nguyen-xu-2025-reasoning,
  title = {Reasoning for Translation: Comparative Analysis of Chain-of-Thought and Tree-of-Thought Prompting for {LLM} Translation},
  author = {Nguyen, Lam and Xu, Yang},
  booktitle = {Proceedings of the 63rd Annual Meeting of the Association for Computational Linguistics (Volume 4: Student Research Workshop)},
  year = {2025},
  address = {Vienna, Austria},
  publisher = {Association for Computational Linguistics},
  url = {https://aclanthology.org/2025.acl-srw.17/},
  doi = {10.18653/v1/2025.acl-srw.17},
  pages = {259--275}
}

@inproceedings{tan-etal-2024-narrowing,
  title = {Narrowing the Gap between Zero- and Few-shot Machine Translation by Matching Styles},
  author = {Tan, Weiting and Xu, Haoran and Shen, Lingfeng and Li, Shuyue Stella and Murray, Kenton and Koehn, Philipp and Van Durme, Benjamin and Chen, Yunmo},
  booktitle = {Findings of the Association for Computational Linguistics: NAACL 2024},
  year = {2024},
  address = {Mexico City, Mexico},
  publisher = {Association for Computational Linguistics},
  url = {https://aclanthology.org/2024.findings-naacl.33/},
  doi = {10.18653/v1/2024.findings-naacl.33},
  pages = {490--502}
}

@inproceedings{agrawal-etal-2023-context,
  title = {In-context Examples Selection for Machine Translation},
  author = {Agrawal, Sweta and Zhou, Chunting and Lewis, Mike and Zettlemoyer, Luke and Ghazvininejad, Marjan},
  booktitle = {Findings of the Association for Computational Linguistics: ACL 2023},
  year = {2023},
  address = {Toronto, Canada},
  publisher = {Association for Computational Linguistics},
  url = {https://aclanthology.org/2023.findings-acl.564/},
  doi = {10.18653/v1/2023.findings-acl.564},
  pages = {8857--8873}
}

@inproceedings{lyu-etal-2025-unveiling,
    title = "Unveiling the Power of Source: Source-based Minimum {B}ayes Risk Decoding for Neural Machine Translation",
    author = "Lyu, Boxuan  and
      Kamigaito, Hidetaka  and
      Funakoshi, Kotaro  and
      Okumura, Manabu",
    booktitle = "Proceedings of the 63rd Annual Meeting of the Association for Computational Linguistics (Volume 1: Long Papers)",
    year = "2025",
    address = "Vienna, Austria",
    publisher = "Association for Computational Linguistics",
    url = "https://aclanthology.org/2025.acl-long.149/",
    doi = "10.18653/v1/2025.acl-long.149",
    pages = "2976--2994"
}

@misc{zhang-etal-2022-rmbr,
    title = "{RMBR}: A Regularized Minimum {B}ayes Risk Reranking Framework for Machine Translation",
    author = "Zhang, Yidan and Wan, Yu and Liu, Dayiheng and Yang, Baosong and He, Zhenan",
    year = "2022",
    eprint = "2203.00201",
    archivePrefix = "arXiv",
    url = "https://arxiv.org/abs/2203.00201"
}

@inproceedings{trabelsi-etal-2024-efficient,
    title = "Efficient Minimum {B}ayes Risk Decoding using Low-Rank Matrix Completion Algorithms",
    author = "Trabelsi, Firas and Vilar, David and Finkelstein, Mara and Freitag, Markus",
    booktitle = "Advances in Neural Information Processing Systems",
    volume = "37",
    pages = "54714--54733",
    year = "2024",
    publisher = "Curran Associates, Inc.",
    url = "https://proceedings.neurips.cc/paper_files/paper/2024/hash/626ab938fe19200324b368f5ee816868-Abstract-Conference.html",
    doi = "10.52202/079017-1734"
}

@misc{bouthors-etal-2025-improving,
    title = "Improving Retrieval-Augmented Neural Machine Translation with Monolingual Data",
    author = "Bouthors, Maxime and Crego, Josep and Yvon, Fran{\c{c}}ois",
    year = "2025",
    eprint = "2504.21747",
    archivePrefix = "arXiv",
    url = "https://arxiv.org/abs/2504.21747"
}

@misc{perrella-etal-2026-span,
    title = "Span-Level Machine Translation Meta-Evaluation",
    author = "Perrella, Stefano and Morales Agostinho, Eric and Zaragoza, Hugo",
    year = "2026",
    eprint = "2603.19921",
    archivePrefix = "arXiv",
    url = "https://arxiv.org/abs/2603.19921"
}

@inproceedings{rei-etal-2022-comet,
    title = "{COMET}-22: Unbabel-{IST} 2022 Submission for the Metrics Shared Task",
    author = "Rei, Ricardo and C. de Souza, Jos{\'e} G. and Alves, Duarte and Zerva, Chrysoula and Farinha, Ana C and Glushkova, Taisiya and Lavie, Alon and Coheur, Luisa and Martins, Andr{\'e} F. T.",
    booktitle = "Proceedings of the Seventh Conference on Machine Translation (WMT)",
    month = dec,
    year = "2022",
    address = "Abu Dhabi, United Arab Emirates (Hybrid)",
    publisher = "Association for Computational Linguistics",
    url = "https://aclanthology.org/2022.wmt-1.52/",
    doi = "10.18653/v1/2022.wmt-1.52",
    pages = "578--585"
}

@inproceedings{heineman-etal-2024-improving,
    title = "Improving Minimum {B}ayes Risk Decoding with Multi-Prompt",
    author = "Heineman, David  and
      Dou, Yao  and
      Xu, Wei",
    booktitle = "Proceedings of the 2024 Conference on Empirical Methods in Natural Language Processing",
    year = "2024",
    address = "Miami, Florida, USA",
    publisher = "Association for Computational Linguistics",
    url = "https://aclanthology.org/2024.emnlp-main.1255/",
    doi = "10.18653/v1/2024.emnlp-main.1255",
    pages = "22525--22545"
}

@inproceedings{alali-issam-2026-maastricht,
    title = "{Maastricht University} at {AMIYA}: Adapting {LLM}s for Dialectal {A}rabic using Fine-tuning and {MBR} Decoding",
    author = "Alali, Abdulhai  and
      Issam, Abderrahmane",
    booktitle = "Proceedings of the 13th Workshop on {NLP} for Similar Languages, Varieties and Dialects",
    year = "2026",
    address = "Rabat, Morocco",
    publisher = "Association for Computational Linguistics",
    url = "https://aclanthology.org/2026.vardial-1.28/",
    doi = "10.18653/v1/2026.vardial-1.28",
    pages = "352--358"
}

@inproceedings{post-2018-call,
    title = "A Call for Clarity in Reporting {BLEU} Scores",
    author = "Post, Matt",
    booktitle = "Proceedings of the Third Conference on Machine Translation: Research Papers",
    year = "2018",
    address = "Brussels, Belgium",
    publisher = "Association for Computational Linguistics",
    url = "https://aclanthology.org/W18-6319/",
    doi = "10.18653/v1/W18-6319",
    pages = "186--191"
}

@inproceedings{hu-etal-2021-lora,
    title = "{L}o{RA}: Low-Rank Adaptation of Large Language Models",
    author = "Hu, Edward J. and Shen, Yelong and Wallis, Phillip and Allen-Zhu, Zeyuan and
      Li, Yuanzhi and Wang, Shean and Wang, Lu and Chen, Weizhu",
    booktitle = "International Conference on Learning Representations",
    year = "2022",
    url = "https://openreview.net/forum?id=nZeVKeeFYf9"
}

@inproceedings{dettmers-etal-2023-qlora,
    title = "{QL}o{RA}: Efficient Finetuning of Quantized {LLM}s",
    author = "Dettmers, Tim and Pagnoni, Artidoro and Holtzman, Ari and Zettlemoyer, Luke",
    booktitle = "Advances in Neural Information Processing Systems",
    volume = "36",
    pages = "10088--10115",
    year = "2023",
    publisher = "Curran Associates, Inc.",
    url = "https://proceedings.neurips.cc/paper_files/paper/2023/hash/1feb87871436031bdc0f2beaa62a049b-Abstract-Conference.html",
    doi = "10.52202/075280-0441"
}

@inproceedings{popovic-2017-chrf,
    title = "chr{F}++: words helping character n-grams",
    author = "Popovi{\'c}, Maja",
    booktitle = "Proceedings of the Second Conference on Machine Translation",
    month = sep,
    year = "2017",
    address = "Copenhagen, Denmark",
    publisher = "Association for Computational Linguistics",
    url = "https://aclanthology.org/W17-4770/",
    doi = "10.18653/v1/W17-4770",
    pages = "612--618"
}

@misc{cheng2025seedx,
    title = "{Seed-X}: Building Strong Multilingual Translation {LLM} with {7B} Parameters",
    author = "Cheng, Shanbo and Bao, Yu and Cao, Qian and Huang, Luyang and Kang, Liyan and Liu, Zhicheng and Lu, Yu and Zhu, Wenhao and Chen, Jingwen and Huang, Zhichao and Li, Tao and Li, Yifu and Lin, Huiying and Liu, Sitong and Peng, Ningxin and She, Shuaijie and Xu, Lu and Xu, Nuo and Yang, Sen and Yu, Runsheng and Yu, Yiming and Zou, Liehao and Li, Hang and Lu, Lu and Wang, Yuxuan and Wu, Yonghui",
    year = "2025",
    eprint = "2507.13618",
    archivePrefix = "arXiv",
    url = "https://arxiv.org/abs/2507.13618"
}

@misc{nllbteam2022nllb,
    title = "No Language Left Behind: Scaling Human-Centered Machine Translation",
    author = "{NLLB Team} and Costa-juss{\`a}, Marta R. and Cross, James and {\c{C}}elebi, Onur and Elbayad, Maha and Heafield, Kenneth and Heffernan, Kevin and Kalbassi, Elahe and Lam, Janice and Licht, Daniel and Maillard, Jean and Sun, Anna and Wang, Skyler and Wenzek, Guillaume and Youngblood, Al and Akula, Bapi and Barrault, Loic and Mejia Gonzalez, Gabriel and Hansanti, Prangthip and Hoffman, John and Jarrett, Semarley and Sadagopan, Kaushik Ram and Rowe, Dirk and Spruit, Shannon and Tran, Chau and Andrews, Pierre and Ayan, Necip Fazil and Bhosale, Shruti and Edunov, Sergey and Fan, Angela and Gao, Cynthia and Goswami, Vedanuj and Guzm{\'a}n, Francisco and Koehn, Philipp and Mourachko, Alexandre and Ropers, Christophe and Saleem, Safiyyah and Schwenk, Holger and Wang, Jeff",
    year = "2022",
    eprint = "2207.04672",
    archivePrefix = "arXiv",
    url = "https://arxiv.org/abs/2207.04672"
}

\clearpage
\appendix

\section{Implementation and Reproducibility Details}
\label{app:repro}

\paragraph{Subtask 1 generators.}
The 12B stage is a QLoRA \citep{dettmers-etal-2023-qlora} SFT of \texttt{gemma-4-12b-it} on 97{,}672 serialized rows (train, AraBench LY/SD with 2,500 and 2,000 rows respectively, public test, and full development): rank~16, alpha~16, dropout~0, learning rate
$2{\times}10^{-4}$ with linear decay, one epoch (3{,}053 steps) at effective batch 32
($8{\times}4$), country-balance temperature $\tau{=}0.5$, seed 3407, maximum sequence length 1024,
and English-only context serialization; CPO follows as in Table~\ref{tab:repro}. The E4B candidate is a
bf16 LoRA SFT of \texttt{gemma-4-E4B-it} on the roughly 85.4k-row train+AraBench+public-test mixture:
rank~32, alpha~64, dropout~0.05, learning rate $2{\times}10^{-4}$, two epochs at effective batch 64 ($4{\times}16$), $\tau{=}1.0$, seed 42, maximum length 1024, decoded
with beam~8 (length penalty 1.0, bf16) over rows with metadata-derived subdialect labels.

The post-evaluation full-data seed and weight-averaging experiment uses a separate recipe from the submitted soup; Appendix~\ref{app:st1constrained} reports its training, merging, and decoding settings alongside all three individual results.

The reference-only Gemma-4-E2B-IT uses the E4B training recipe and data mixture above, with beam-8 decoding. Together with the six API model-prompt variants below, it forms the seven-member jury.

\paragraph{Inference-only components.}
The six reference-only
jurors are \texttt{gpt-4.1-mini} (subdialect-named prompt; temperature 0, 300 output tokens),
\texttt{gpt-5.5} (subdialect-named and full-metadata prompts; reasoning effort none, 800 output
tokens), \texttt{gemini-3.6-flash} (subdialect-named), and \texttt{gemini-3.1-pro-preview}
(subdialect-named and full-metadata), the Gemini pair at thinking level low with 512 output
tokens and no sampling controls. The Subtask 2 retriever uses \texttt{intfloat/multilingual-e5-large} with mean pooling; its translation generator is \texttt{unsloth/gemma-4-31b-it}, decoded at temperature 0. The primary Subtask 3 proposer is \texttt{gpt-5.5} with 8{,}000-token completions. \texttt{gemini-3.6-flash}, at temperature 0, supplies spans for extended widening, agreement-gated additions, and the empty-row fallback below.

\begin{table*}[t]
\centering
\small
\setlength{\tabcolsep}{4pt}
\begin{tabular}{>{\raggedright\arraybackslash}p{0.16\textwidth}>{\raggedright\arraybackslash}p{0.19\textwidth}>{\raggedright\arraybackslash}p{0.26\textwidth}>{\raggedright\arraybackslash}p{0.27\textwidth}}
\toprule
System & Data & Adaptation or retrieval & Inference and role \\
\midrule
ST1 unconstrained, 12B CPO & Alexandria train, development, and public test; AraBench (2,500 LY + 2,000 SD examples)
\citep{sajjad-etal-2020-arabench} & Gemma-4-12B-IT; 13,049 CPO pairs; $\beta=0.1$; SFT anchor 1.0;
lr $10^{-5}$; one epoch; effective batch 16; seed 3407 & selectable local candidate; metadata
conditioning; terminal-punctuation normalization \\
ST1 unconstrained, E4B & Alexandria train and public test plus AraBench LY/SD; no development exposure &
Gemma-4-E4B-IT LoRA SFT & beam-8; selectable local candidate \\
ST1 constrained & train+development soup; separate train+public-test adapter routed only for LY/SD & one
NileChat-3B-Base backbone; three rank-32, alpha-64 soup deltas plus one rank-32, alpha-64 fold-in delta &
country-selected LoRA route; derived subdialect label; beam-8; self-context; MR/MA $k{=}3$ demonstrations \\
ST2 & final retrieval uses released blind sources; ArBanking77 used only for exploratory retriever selection;
no parallel task training data & multilingual E5-Large; reciprocal 32-neighbor filtering; margin ranking;
$k=32$ & Gemma-4-31B-IT; greedy decoding; 32k context; guided generation; exact-copy regeneration \\
ST3 & 1,125 training rows; 138 development rows for epoch and system selection & \texttt{UBC-NLP/MARBERTv2} joint 13-label BIO;
direction as first segment; lr $2\times10^{-5}$; batch 16; max length 256; seed 7; up to 25 epochs &
GPT-5.5 proposer; MARBERTv2 boundary control; agreement-gated Gemini 3.6 Flash additions; empty-row rescue;
gap-1 same-category merge \\
\bottomrule
\end{tabular}
\caption{Final-system implementation summary. Selectable ST1 candidates are summarized in Table~\ref{tab:candidates}; reference-only jurors and their prompts are described in Appendix~\ref{app:repro}.}
\label{tab:repro}
\end{table*}

\paragraph{Subtask 3 empty-row fallback.}
For any row still empty after extended widening and agreement-gated additions, the submitted controller tries, in order: (1) the Gemma-4-31B-IT few-shot extractor combined with MARBERTv2; (2) Gemini 3.6 Flash alone; (3) the supervised-fine-tuned Gemma-4-12B-IT extractor combined with MARBERTv2; and (4) MARBERTv2 alone. Both local hybrids use span union with widest-span merging of overlapping predictions. The controller takes the first nonempty span set, then merges adjacent same-category fragments.

\subsection{Prompt contracts}

The following schemas record the prompt choices tied to measured system behavior. Bracketed text
denotes inserted data; provider-specific request parameters are reported above.

\paragraph{Subtask 1 reference-only jurors.}
Four model-prompt variants, \texttt{gpt-4.1-mini}, \texttt{gpt-5.5},
\texttt{gemini-3.6-flash}, and \texttt{gemini-3.1-pro-preview}, use the same
subdialect-named textual contract through provider-specific message wrappers. Its system instruction is:
\begin{quote}
\small\raggedright
\texttt{You are a professional, precise, and native-level translator. Your task is to translate the provided text from the source language into the specified target language. Output only the best possible translation, with no explanations, alternatives, or additional text. Preserve formatting, punctuation, and any untranslatable elements (such as names or technical terms) as appropriate.}
\end{quote}
The user message is:
\begin{quote}
\small\raggedright
\texttt{Translate the following text into [derived subdialect], only print out the translation,
not add any explanations: [English turn]}
\end{quote}
The \texttt{gpt-5.5} and \texttt{gemini-3.1-pro-preview} full-metadata variants additionally expose
the same released fields, with minor serializer-specific wording and formatting. The GPT-5.5 full-metadata system instruction is:
\begin{quote}
\small\raggedright
\texttt{You are a native-speaker translator who produces authentic spoken dialectal Arabic (NOT Modern Standard Arabic). Match the exact local dialect requested. Preserve speaker/addressee gender agreement. Keep conventional code-switched loanwords, names, numbers, URLs, and emojis as-is. Output ONLY the translation, no quotes, no explanations, no transliteration.}
\end{quote}
The Gemini full-metadata variant uses this shorter instruction, concatenated before the user text rather than sent as a separate system message:
\begin{quote}
\small\raggedright
\texttt{You are a native-speaker translator who produces authentic spoken dialectal Arabic (NOT Modern Standard Arabic). Match the exact local dialect requested. Preserve speaker/addressee gender agreement. Keep code-switched loanwords, names, numbers as-is. Output ONLY the translation.}
\end{quote}
The full-metadata user-message schema is:
\begin{quote}
\small\raggedright
\texttt{Target dialect: [derived subdialect] ([country]).}\\
\texttt{Domain: [domain].}\\
\texttt{Speaker: [speaker] -> Addressee: [addressee].}\\
\texttt{Gender direction (speaker -> addressee): [direction].}\\
\texttt{Prior dialogue turns: [preceding English turns].}\\
\texttt{Now translate ONLY this turn into authentic [derived subdialect]: [English turn]}
\end{quote}
No reference Arabic history is supplied. The Gemini metadata serializer retains an empty country-tagged Arabic line after each preceding English turn; GPT-5.5 includes only the English history lines. Conversation identifiers are not inserted directly; they are used only to derive the subdialect label.

\paragraph{Subtask 1 constrained completion.}
Training and inference use the same completion frame:
\begin{quote}
\small\raggedright
\texttt{You are an expert translator. Translate the English sentence into [derived subdialect].}\\
\texttt{\#\#\# Metadata: [country, domain, participants, speaker, direction]}\\
\texttt{\#\#\# Conversation History: [preceding English turns and Arabic translations]}\\
\texttt{\#\#\# Sentence to Translate: [English turn]}\\
\texttt{\#\#\# Translation:}
\end{quote}
Each training example is one English turn in this frame, with the preceding turns of its
conversation serialized as history; during inference, the Arabic history contains the model's own
preceding outputs. For MR and MA,
three complete training instances serialized in this frame precede the query, preserving the
format used during fine-tuning.

\paragraph{Subtask 2 retrieval-guided generation.}
The prompt names the source and target dialects, inserts the retrieved sentences, and explicitly
defines their role:
\begin{quote}
\small\raggedright
\texttt{The following are authentic [target] banking usage examples retrieved only from the
supplied source sentences. They are not translations of the input. Use only their [target]
vocabulary, morphology, spelling, and style. Do not copy their meaning or transaction details.}
\end{quote}
The retained generation instruction then asks the model to consider internally (1) a conservative
meaning-preserving rendering, (2) a natural target-dialect rendering, and (3) a rendering aligned
with the retrieved usage. It returns only one synthesized sentence, without alternatives, analysis,
or scores.

\paragraph{Subtask 3 tagged span annotation.}
Four training examples precede the query, ranked first by matching translation direction and then
by Arabic token overlap; at least one example covers a rare LQM category. The proposer then
receives the direction, English source, and Arabic MT, followed by
\texttt{Tagged:}. Its output contract is to copy the Arabic MT verbatim and wrap each minimal error
span as \texttt{<category>span</category>}, where \texttt{category} is one of the six LQM labels.
The instruction resolves the dominant ambiguity by assigning a meaning-preserving variety or
register mismatch to sociolinguistics, and a meaning change to semantics. Only the tagged Arabic
copy is returned. We extract the tagged substrings and locate them in the original MT, so offsets
are reconstructed rather than generated numerically.

\FloatBarrier
\section{Official Score Details}
\label{app:official}

Table~\ref{tab:results} reports the aggregate blind-test results. For Subtask 3, Overall is the mean
of overlap-span F1 (0.5426) and pooled error-class F1 (0.4538); exact-span F1 is 0.2319 and does not enter Overall. For reproducibility, we follow the deployed scorer, whose
class component pools overlap-matched spans before computing F1. Our development implementation
reproduces the deployed score.

Score differences are computed before rounding whenever full-precision scorer records are available; the Subtask 3 comparison record uses the preserved four-decimal scores.

Table~\ref{tab:st1percountry} reports the two officially submitted Subtask 1 systems by country. The unconstrained system scores higher in all 13 countries and on both metrics. Both systems score highest on Syrian Arabic and lowest on Mauritanian Arabic.

\begin{table}[t]
\centering
\small
\setlength{\tabcolsep}{2.5pt}
\begin{tabular}{lrrrr}
\toprule
& \multicolumn{2}{c}{Constrained\textsuperscript{*}} & \multicolumn{2}{c}{Unconstrained\textsuperscript{\dag}} \\
\cmidrule(lr){2-3} \cmidrule(lr){4-5}
Country & spBLEU $\uparrow$ & chrF++ $\uparrow$ & spBLEU $\uparrow$ & chrF++ $\uparrow$ \\
\midrule
EG & 33.07 & 47.36 & 37.54 & 51.37 \\
JO & 36.17 & 50.01 & 38.80 & 52.79 \\
LB & 32.44 & 46.97 & 35.75 & 49.56 \\
LY & 26.72 & 43.43 & 29.45 & 46.17 \\
MA & 24.14 & 40.57 & 27.57 & 42.90 \\
MR & 17.67 & 34.83 & 20.18 & 36.58 \\
OM & 35.51 & 49.41 & 38.10 & 51.57 \\
PS & 33.22 & 47.64 & 36.04 & 50.15 \\
SA & 32.86 & 47.47 & 36.95 & 51.06 \\
SD & 28.17 & 43.10 & 30.32 & 45.25 \\
SY & 39.75 & 53.77 & 42.92 & 56.52 \\
TN & 29.41 & 43.55 & 33.16 & 47.16 \\
YE & 26.37 & 42.42 & 28.57 & 44.64 \\
\midrule
\textbf{Macro} & 30.4233 & 45.4250 & 33.4891 & 48.1320 \\
\bottomrule
\end{tabular}
\caption{Per-country blind-test scores for our two first-place official Subtask 1 submissions. \textsuperscript{*}Country-routed NileChat (30.4233 macro spBLEU): LY and SD use the train+public-test adapter; the other 11 countries use the train+development soup. \textsuperscript{\dag}Candidate-constrained MBR (33.4891), with the four candidates in Table~\ref{tab:candidates}. Neither column includes post-evaluation changes. Codes: EG Egypt, JO Jordan, LB Lebanon, LY Libya, MA Morocco, MR Mauritania, OM Oman, PS Palestine, SA Saudi Arabia, SD Sudan, SY Syria, TN Tunisia, YE Yemen. Per-country values are rounded to two decimals; macro scores are reported by the scorer.}
\label{tab:st1percountry}
\end{table}

\FloatBarrier
\section{Subtask 1: Extended Experiments and Analysis}
\label{app:st1}

\subsection{Constrained-track ablations}
\label{app:st1constrained}

\begin{table*}[t]
\centering
\small
\setlength{\tabcolsep}{5pt}
\begin{tabular}{lrrrrr}
\toprule
Training mixture & Macro spBLEU & $\Delta$ & LY & SD & Mean, other 11 \\
\midrule
Training + development & 29.2323 & baseline & 20.9637 & 21.4406 & 30.6923 \\
\addlinespace[2pt]
\multicolumn{6}{l}{\emph{Added-data budget matched to the public LY/SD subset}} \\
$\quad$Duplicated training rows & 29.0971 & -0.1352 & 20.8941 & 21.7975 & 30.5064 \\
$\quad$Authentic public LY/SD rows & 30.1179 & +0.8856 & 26.7867 & \textbf{27.8097} & 30.6305 \\
\addlinespace[2pt]
\multicolumn{6}{l}{\emph{Added-data budget matched to the complete public-test split}} \\
$\quad$Duplicated training rows & 28.8483 & -0.3840 & 21.7299 & 21.2975 & 30.1819 \\
$\quad$Authentic complete public test & \textbf{30.4733} & \textbf{+1.2410} & \textbf{27.2953} & 27.5768 & \textbf{31.0256} \\
\bottomrule
\end{tabular}
\caption{Unofficial post-evaluation supervision controls produced by the official scorer on the
blind test. All rows use the same seed-42 adapter recipe and inference configuration. $\Delta$ is
relative to training on the training and development splits. The duplicate rows contain no new
translations. Bold marks the highest value in each numeric column.}
\label{tab:st1data}
\end{table*}

\begin{figure*}[t]
\centering
\includegraphics[width=0.94\textwidth]{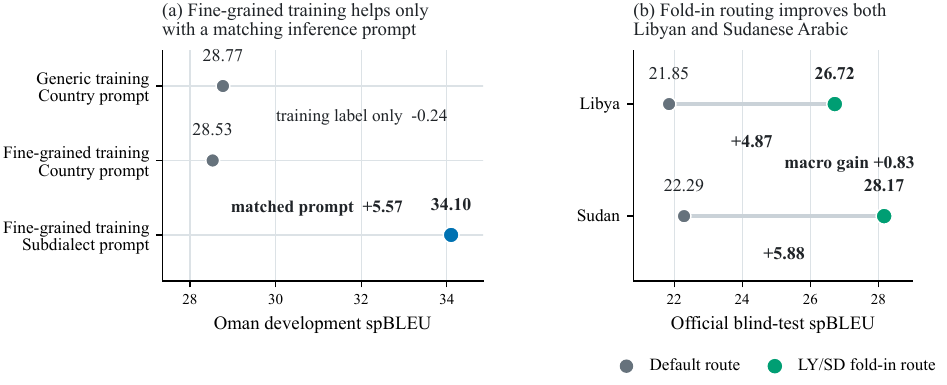}
\caption{Subtask 1 conditioning and constrained routing controls. Panel (a) compares two Omani-only
\texttt{Gemma-4-12B-IT} LoRA adapters trained on 6,280 rows and evaluated on 1,109 held-out Omani
development turns. The adapters share the base model, recipe, and seed; only the country-level
versus subdialect-level training label changes. The subdialect-trained adapter is evaluated under
both country-level and matched subdialect prompts. Panel (b) shows the official blind-test changes
from routing LY/SD to the fold-in adapter.}
\label{fig:conditioning}
\end{figure*}

\paragraph{Authentic supervision versus additional updates.}
Table~\ref{tab:st1data} compares five seed-42 \texttt{NileChat-3B-Base} adapters trained with the
same LoRA recipe and decoded identically. At two added-data budgets, duplicating existing rows
controls for both the number of examples and optimizer updates. Authentic public-test supervision
outperforms its matched duplicate-row control by 1.0208 macro spBLEU when only LY/SD rows are added
and by 1.6250 when the complete public-test split is added. The gain therefore follows new labeled
content rather than a longer optimization schedule.

\paragraph{Seed variation and weight averaging.}
We repeat the full-data recipe with seeds 42, 123, and 7 (Table~\ref{tab:st1seeds}). Each \texttt{NileChat-3B-Base} run uses all 93,172 training, development, and public-test examples: LoRA rank 32, alpha 64, dropout 0.05, attention and MLP projections, AdamW at $2\times10^{-4}$, bf16, two epochs, and effective batch 16 (11,648 optimizer steps). Only the random and data-order seed changes. Training uses gold dialogue history. At inference, all arms use beam-8, derived subdialect labels, and generated Arabic self-context, without a special LY/SD hint.

Following weight-space averaging \citep{pmlr-v162-wortsman22a}, we merge each LoRA adapter into the same backbone, average the three complete models uniformly in fp32, and save one bf16 model. We do not average the LoRA factors directly or combine predictions at inference. This requires three training runs but retains one 3.086B-parameter model for decoding. The two additional seeds and equal-weight soup were fixed before their post-evaluation scores were obtained.

The individual seeds have mean macro spBLEU 30.4537 and sample standard deviation 0.0171. Their soup reaches 30.9726, improving on the mean by 0.5189 and on the best seed by 0.4993. It exceeds the strongest seed within each of the 13 dialects, with margins from 0.0131 in YE to 0.6433 in MR. Macro chrF++ also improves, by 0.3884 over the strongest individual model on that metric. These results support weight averaging for this full-data recipe; the three-seed dispersion is descriptive, not a confidence interval or significance test. Appendix~\ref{app:st1jury} reports a separate post-evaluation experiment replacing the submitted unconstrained system's NileChat candidate with this soup.

\begin{table}[t]
\centering\small
\begin{tabular}{lrr}
\toprule
Full-data model & Macro spBLEU & Macro chrF++ \\
\midrule
Seed 42 & 30.4733 & 45.4750 \\
Seed 123 & 30.4426 & 45.4991 \\
Seed 7 & 30.4451 & 45.4739 \\
\midrule
Equal-weight soup & \textbf{30.9726} & \textbf{45.8875} \\
\bottomrule
\end{tabular}
\caption{Post-evaluation full-data seed comparison, using the official scorer on the blind test. The three seeds share one training recipe; all four arms use beam search with width 8 and sampling disabled. The soup averages the three merged models with equal weights and is decoded independently. It is not the train+development soup used in the submitted system. Bold identifies the soup.}
\label{tab:st1seeds}
\end{table}

\paragraph{Decoding.}
For the full-data seed-42 adapter, greedy decoding scores 26.6855, beam-4 scores 30.3822, and beam-8 scores 30.4733. Beam search supplies nearly all of the 3.7878-point gain over greedy decoding, while increasing the width from four to eight adds 0.0912. On the 11-country development set, beam-8 exceeds beam-5 by 0.121 macro spBLEU, with a 95\% conversation-bootstrap interval of $[+0.025,+0.222]$.

\paragraph{Fine-grained labels and generated dialogue context.}
Table~\ref{tab:st1context} isolates two inference choices in one fixed system: the seed-42 \texttt{NileChat-3B-Base} LoRA adapter trained on all released training, development, and public-test data and decoded in bf16 with beam width 8. First, the prompt names either the released country or the subdialect derived from the conversation-prefix mapping in Section~\ref{sec:subdialect}. Second, target-side dialogue history is either omitted or generated by the model itself. Every arm retains the preceding English turns. With generated self-context, the conversation is translated in order and the model's own preceding Arabic translations, never reference translations, are supplied when translating the next turn. Averaged over the other factor, the derived label contributes 1.6136 macro spBLEU, generated self-context contributes 0.1707, and their interaction, the difference between the two simple effects, is 0.0478. Their effects are close to additive in this fixed system.

\begin{table}[t]
\centering
\small
\begin{tabular}{lrr}
\toprule
Prompt label & \shortstack{English history\\only} & \shortstack{+ Generated\\Arabic history} \\
\midrule
Country-level & 28.6890 & 28.8358 \\
Derived subdialect & 30.2787 & \textbf{30.4733} \\
\bottomrule
\end{tabular}
\caption{Unofficial post-evaluation factorial for the full-data seed-42 NileChat-3B-Base adapter, scored on the official blind test. All arms retain English source history; Arabic history contains prior model outputs only. Bold marks the best configuration.}
\label{tab:st1context}
\end{table}

\paragraph{Why label granularity matters.}
Figure~\ref{fig:conditioning}(a) isolates the interaction between training-label and test-prompt
granularity. We fine-tune two Omani-only \texttt{Gemma-4-12B-IT} LoRA adapters on the same 6,280
training rows and evaluate them on the same 1,109 held-out Omani development turns. The base model,
LoRA recipe, and seed are identical; only the training prompt differs. One adapter is trained with
the generic label \emph{Omani Arabic}, while the other receives the annotated Al-Wafi, Rustaqi, or
Suri label for each example. At inference, the generic adapter with a country prompt scores 28.77.
The subdialect-trained adapter scores 28.53 with that same coarse prompt, but 34.10 when given the
matching subdialect, a gain of 5.57. Fine-grained supervision therefore helps only when its label
granularity is preserved at inference.

The controlled result explains the broader near-matched official comparison in Section~\ref{sec:subdialect}. The five countries containing multiple annotated subdialects contribute +1.14 to the 13-country macro average (OM +7.47, SY +2.73, YE +2.13, PS +1.43, and SA +1.01); the other eight contribute only +0.04. Lexical analysis suggests a plausible mechanism: Omani subdialects, for example, use different forms for common functions such as \emph{want}. Across the five multi-subdialect countries, lexical separability and the official gain have a positive exploratory correlation ($r=0.75$); the sample is too small for a general claim.

Figure~\ref{fig:conditioning}(b) shows the complementary data intervention. Routing LY and SD to the
fold-in adapter raises their official spBLEU by 4.87 and 5.88, respectively, while the other 11
country outputs remain byte-identical.

\subsection{Model landscape and translation behavior}

\paragraph{Model landscape.}
Table~\ref{tab:st1panel} summarizes our broader model study. The first two blocks share one common full-development evaluation; the remaining blocks record exploratory screens or dialect-specific slices. On the common evaluation, Gemma-4-E4B-IT rises from 23.49 zero-shot to 33.58 after fold-in SFT and beam search. The corresponding official-test system scores 32.2336, making the compact E4B our strongest standalone submitted candidate. Excluding the daggered training-overlap diagnostic, it also exceeds every tested 12B and 31B configuration in the common evaluation, showing that raw parameter count alone does not determine performance here.

\begin{table*}[p]
\centering
\small
\setlength{\tabcolsep}{4pt}
\begin{tabular}{lllr}
\toprule
Model & Adaptation and decoding & Evaluation set & spBLEU \\
\midrule
\multicolumn{4}{l}{\emph{Common full-development evaluation: prompted systems, 12,250 turns, 11 countries}} \\
\midrule
Gemma-3-1B-IT & zero-shot, greedy & full development & 3.77 \\
Gemma-3-4B-IT & zero-shot, greedy & full development & 10.50 \\
NileChat-3B (untuned) & zero-shot, greedy & full development & 13.11 \\
Gemma-4-E2B-IT & zero-shot, greedy & full development & 21.10 \\
Gemma-4-E4B-IT & zero-shot, greedy & full development & 23.49 \\
Gemini 2.5 Pro & zero-shot, prompt-matched & full development & 23.93 \\
Gemini 3.6 Flash & zero-shot, prompt-matched & full development & 24.04 \\
Gemma-4-31B-IT & zero-shot, greedy & full development & 24.93 \\
Gemma-4-31B-IT & BM25 few-shot, $k{=}16$ & full development & 27.93 \\
Gemma-4-31B-IT & identifier-group BM25 few-shot, $k{=}32$ & full development & 30.07 \\
\midrule
\multicolumn{4}{l}{\emph{Common full-development evaluation (continued): fine-tuned systems}} \\
\midrule
Gemma-3-4B (base) & LoRA SFT, beam-8 & full development & 29.60 \\
Gemma-3-4B-IT & LoRA SFT, beam-8 & full development & 29.82 \\
NileChat-3B-Base & constrained LoRA SFT, beam-8 & full development & 29.86 \\
Qwen3.5-9B & fold-in LoRA SFT, beam-5 & full development & 30.72 \\
Gemma-4-E2B-IT & LoRA SFT, beam-8 & full development & 31.71 \\
Gemma-4-12B-IT & LoRA SFT, train only, greedy & full development & 31.95 \\
Gemma-4-12B-IT & + LY/SD fold-in and DPO, greedy & full development & 32.06 \\
Gemma-4-E2B-IT & fold-in SFT, beam-8 & full development & 32.20 \\
Gemma-4-12B-IT & + CPO, MR/TN routing, greedy & full development & 32.63 \\
Gemma-4-E4B-IT & LoRA SFT, beam-8 & full development & 32.99 \\
Gemma-4-E4B-IT & fold-in SFT, beam-8 & full development & \textbf{33.58} \\
Gemma-4-12B-IT & fold-in SFT + CPO & development used in training & 34.13$^\dagger$ \\
\midrule
\multicolumn{4}{l}{\emph{Exploratory samples, screens, and holdout folds}} \\
\midrule
Hy-MT2-1.8B & zero-shot, greedy & 110-turn sample & 9.62 \\
Granite-4.0-H-3B-AR & zero-shot, greedy & 55-turn sample & 12.68 \\
Qwen3.5-9B & zero-shot, greedy & 440-turn sample & 13.25 \\
Command-R7B-Arabic & zero-shot, greedy & 440-turn sample & 16.91 \\
GPT-5 & zero-shot & 440-turn sample & 19.85 \\
Seed-X-PPO-7B & LoRA SFT, 0.36 epoch, beam-3 & 440-turn sample & 21.40 \\
Gemma-4-12B-IT & zero-shot, greedy & 1,100-turn screen & 24.04 \\
Qwen3.5-9B & LoRA SFT, beam-5 & 2,442-turn holdout fold & 31.18 \\
\midrule
\multicolumn{4}{l}{\emph{Dialect-specific slices of the development set}} \\
\midrule
DarijaTranslation-V1 & off the shelf & MA column & 5.63 \\
Masrawy v2 (oddadmix) & off the shelf & EG column & 13.77 \\
NLLB-200-1.3B & MA-only SFT & MA column & 14.79 \\
Atlas-Chat-2B & MA-only LoRA SFT & MA column & 14.98 \\
Masrawy v2.9 (NAMAA) & off the shelf & EG column & 15.14 \\
NLLB + AraDiCE & two-stage EN to MSA to dialect & 5 columns, 200 turns each & 16.23 \\
50M-Egyptian-v1 & off the shelf, own template & EG column & 18.28 \\
English2Dialect & off the shelf & 5-column mean & 20.03 \\
Levanti & off the shelf & 4-column mean & 23.86 \\
\bottomrule
\end{tabular}
\caption{Subtask 1 model landscape, separated by evaluation protocol. Rows in the first two common full-development blocks are directly comparable except for the daggered diagnostic, which includes the development set in training. Other blocks report only the stated sample, fold, or dialect slice. Here, fold-in denotes an expanded mixture that adds released public-test supervision and, where used by the experiment, AraBench LY/SD rows. Bold marks the strongest held-out full-development result. DPO: direct preference optimization.}
\label{tab:st1panel}
\end{table*}

The $k{=}32$ Gemma-4-31B-IT row first restricts demonstrations to the query's released
conversation-identifier group and then uses BM25 to fill any remaining slots. It is included to
document the tested retrieval strategy, not as an independent training recipe. The daggered CPO
row is likewise a diagnostic because its evaluation data were included in its training mixture.
The NLLB and Seed-X-PPO model families are described by \citet{nllbteam2022nllb} and
\citet{cheng2025seedx}, respectively.

\paragraph{What remains beyond mechanical normalization.}
One simple surface intervention has a measurable but limited effect: on the 12B CPO candidate, deterministic terminal-punctuation normalization raises official blind-test macro spBLEU from 31.7873 to 32.0318, a gain of 0.2446. Separately, Table~\ref{tab:st1errors} analyzes a 12B CPO checkpoint preceding the submitted candidate on the 12,250-turn development set covering 11 countries, using a layered word-level taxonomy. Exact matches account for 50.0\% of reference words, while spelling and clitic-level morphology together account for under 10\%. The largest non-exact-match category, at 39.2\%, contains different surface forms after normalization that string matching alone cannot separate into valid paraphrases and mistranslations.

Additional probes find little recoverable mass in simple transformations. Dialect-cluster leakage
accounts for 0.2\%; the prediction-to-reference length ratio is 1.06, giving no evidence of
systematic shortening; and all numeric tokens form only 0.17\% of reference words, with an oracle ceiling of
+0.03 macro spBLEU. Normalizing both predictions and references for spelling and morphology changes macro spBLEU by +1.59, with gains in MR (+12.6) and YE (+11.3) offset by losses including EG ($-7.2$) and MA ($-6.1$). This diagnostic changes the scoring inputs on both sides; it is not an achievable post-processing gain. Manual inspection of the MR cases reveals many synonym
pairs, such as \textit{l\=azim} and \textit{\d{d}ar\=ur\=i} for ``must'' or ``necessary.'' For MR, the roughly 30\% exact-match and 60\% different-word profile is consistent with substantial single-reference
variation, but source-grounded native-speaker annotation would be required to distinguish that
variation from genuine translation errors.

\begin{table}[t]
\centering
\small
\begin{tabular}{p{0.53\columnwidth}r}
\toprule
Alignment outcome & Reference words (\%) \\
\midrule
Exact match & 50.0 \\
Spelling variant & 1.8 \\
Clitic-level morphology & 7.9 \\
Different surface form after normalization & 39.2 \\
Code-switch & $\sim$1.0 \\
Unmatched numeral & $\sim$0.1 \\
\bottomrule
\end{tabular}
\caption{Word-level alignment outcomes for the earlier 12B CPO development checkpoint after layered normalization. Percentages use all reference words as the denominator; code-switch and numeral rows count only words still unmatched. The morphology class contains a matching stem with a different clitic.}
\label{tab:st1errors}
\end{table}

\FloatBarrier
\subsection{Selection granularity and utility}

\paragraph{Selection granularity.}
The submitted selector changes systems on 71.4\% of consecutive turns. A switching penalty of
$\lambda=2$ lowers the rate to 54.1\% but costs 0.0850 macro spBLEU. Selecting one system for an
entire conversation costs 0.7391, although it remains 0.5163 above the best standalone candidate.
Under single-reference spBLEU, turn-level complementarity is therefore more valuable than enforcing
system consistency across a conversation.

\begin{table}[t]
\centering
\small
\setlength{\tabcolsep}{4pt}
\begin{tabular}{lrrr}
\toprule
Pairwise utility & Changed (\%) & spBLEU & chrF++ \\
\midrule
Sentence-spBLEU & baseline & \textbf{33.4891} & 48.1320 \\
Sentence-chrF++ & 23.4 & 33.0389 & \textbf{48.3006} \\
COMET-22 & 60.2 & 32.0266 & 47.0838 \\
\bottomrule
\end{tabular}
\caption{Utility comparison under a fixed candidate pool and output-eligibility rule. Changed
selections are measured against the submitted sentence-spBLEU utility. The alternatives are
unofficial post-evaluation controls. Bold marks the best score for each metric.}
\label{tab:st1utility}
\end{table}

\paragraph{Utility alignment.}
Table~\ref{tab:st1utility} freezes the pool, roles, deduplication, and restricted argmax while
changing only the pairwise utility. Sentence-chrF++ improves official chrF++ by 0.1686 but gives up
0.4501 spBLEU. COMET-22 \citep{rei-etal-2022-comet} changes 60.2\% of selections, largely toward the NileChat soup, and falls
below the strongest standalone candidate on both official metrics. Under this fixed pool, each
string utility favors its corresponding metric, while the COMET-22 utility improves neither
official string metric.
This experiment complements semantic, efficient, and source-based MBR variants
\citep{zhang-etal-2022-rmbr,trabelsi-etal-2024-efficient,lyu-etal-2025-unveiling}.

\subsection{Candidate, jury, and eligibility contributions}
\label{app:st1jury}

\paragraph{Jury-family interaction.}
The two official submissions and two post-evaluation controls form a complete $2\times2$ factorial
over the open E2B juror and the six API jurors (Table~\ref{tab:st1juryfactorial}). E2B contributes
0.1025 without the API jurors and 0.1703 with them; the API family contributes 0.2263 without E2B
and 0.2940 with it. Their positive interaction, computed with the same difference-of-differences convention, is 0.0677, and the API family improves 10 of 13
dialects in either E2B condition.

\begin{table}[t]
\centering
\small
\begin{tabular}{lrr}
\toprule
Reference-only jury & No API jurors & Six API jurors \\
\midrule
No E2B & 33.0925 (post) & 33.3188 (post) \\
E2B & 33.1950 & \textbf{33.4891} \\
\bottomrule
\end{tabular}
\caption{Jury-family factorial with the four selectable candidates and fixed sentence-spBLEU
utility. Parentheses mark post-evaluation controls; the other two cells are official submissions. Bold marks the highest score.}
\label{tab:st1juryfactorial}
\end{table}

\paragraph{Juror stability.}
Table~\ref{tab:jurors} compares all eight tested reference-only jurors on two tuning splits. Among the seven retained jurors, only the E2B system, the GPT-5.5 target-dialect prompt, and the Gemini 3.6 target-dialect prompt have positive leave-one-out effects on both splits. Adding the excluded GPT-5.4 mini back to the final pool lowers scores on both splits. Exact development Shapley values over all 256 coalitions of the eight tested jurors sum to that diagnostic pool's +0.4329 improvement over four-candidate MBR. The signs differ across splits and attribution methods, so these analyses diagnose juror behavior rather than define a tuned subset.

\begin{table*}[t]
\centering
\small
\setlength{\tabcolsep}{11pt}
\begin{tabular}{lrrr}
\toprule
Reference-only juror & Development effect & Public-test effect & Development Shapley \\
\midrule
Gemma E2B, fine-tuned                  & +0.020 & +0.771 & -0.018 \\
GPT-4.1 mini, dialect prompt           & +0.161 & -0.025 & +0.086 \\
GPT-5.4 mini, dialect prompt           & -0.068 & -0.086 & -0.062 \\
GPT-5.5, dialect prompt                & +0.144 & +0.018 & +0.086 \\
GPT-5.5, metadata prompt               & +0.063 & -0.042 & +0.093 \\
Gemini 3.6 Flash, dialect prompt       & +0.023 & +0.012 & +0.062 \\
Gemini 3.1 Pro, dialect prompt         & +0.200 & -0.072 & +0.085 \\
Gemini 3.1 Pro, metadata prompt        & +0.156 & -0.031 & +0.101 \\
\bottomrule
\end{tabular}
\caption{Juror effects across splits and attribution methods. For the seven retained jurors, the first two numeric columns report the final-pool score minus the leave-one-out score. The excluded GPT-5.4 mini row instead reports its add-back score minus the final-pool score. Seven of eight development effects are positive, compared with three on public test. The third column gives exact development Shapley values over all 256 coalitions of the eight tested jurors, including GPT-5.4 mini; these sum to the eight-juror diagnostic pool's +0.4329 gain. Positive values indicate a beneficial contribution.}
\label{tab:jurors}
\end{table*}

\paragraph{Candidate contributions.}
In post-evaluation controls, demoting one local candidate to reference-only while retaining its reference hypotheses lowers macro
spBLEU by 0.1997 for 12B CPO, 0.1332 for 12B CPO + ICL, 0.4175 for E4B, and 0.1366 for the
NileChat soup. Each candidate therefore makes a measurable conditional contribution under the
fixed selector, with E4B contributing most strongly and helping all 13 countries.

Table~\ref{tab:st1bydialect} shows a second form of complementarity. The best complete candidate
varies by dialect: ICL leads in seven, E4B in four, and 12B CPO in two. Although the NileChat soup
leads none, it supplies 12.6\% of selected turns. MBR beats every dialect's strongest complete
candidate, from +0.09 in LY to +1.51 in MR, and exceeds a retrospective dialect-level system oracle,
which selects one complete candidate per dialect using blind-test scores, by 0.8145. Its gains
therefore arise within dialects rather than only from coarse routing.

\begin{table*}[t]
\centering
\small
\begin{tabular}{l rrrr @{\hspace{1.6em}} r r}
\toprule
 & \multicolumn{4}{c}{selectable candidates (standalone)} & \multicolumn{2}{c}{selector} \\
\cmidrule(lr){2-5}\cmidrule(lr){6-7}
Dialect & 12B CPO & 12B ICL & E4B & NileChat & MBR & $\Delta$ \\
\midrule
EG & 35.69 & \textbf{36.91} & 35.62 & 33.07 & 37.54 & +0.63 \\
JO & 37.32 & \textbf{38.39} & 37.67 & 36.17 & 38.80 & +0.41 \\
LB & 33.48 & \textbf{34.83} & 33.80 & 32.44 & 35.75 & +0.93 \\
LY & 29.05 & 27.90 & \textbf{29.37} & 21.85 & 29.45 & +0.09 \\
MA & \textbf{26.47} & 26.38 & 25.69 & 24.14 & 27.57 & +1.10 \\
MR & 18.06 & \textbf{18.67} & 18.49 & 17.67 & 20.18 & +1.51 \\
OM & 36.00 & 35.98 & \textbf{37.92} & 35.51 & 38.10 & +0.18 \\
PS & 34.86 & 33.83 & \textbf{35.32} & 33.22 & 36.04 & +0.73 \\
SA & 35.34 & \textbf{35.86} & 35.54 & 32.86 & 36.95 & +1.09 \\
SD & 29.61 & 29.11 & \textbf{29.61} & 22.29 & 30.32 & +0.71 \\
SY & \textbf{41.78} & 41.41 & 41.66 & 39.75 & 42.92 & +1.14 \\
TN & 31.31 & \textbf{32.12} & 30.86 & 29.41 & 33.16 & +1.04 \\
YE & 27.44 & \textbf{27.53} & 27.49 & 26.37 & 28.57 & +1.04 \\
\midrule
Macro & 32.0318 & 32.2235 & \textbf{32.2336} & 29.5969 & 33.4891 & +1.2554 \\
Retrospective dialect-level system oracle & \multicolumn{4}{c}{32.6746} & 33.4891 & +0.8145 \\
\bottomrule
\end{tabular}

\caption{Official blind-test spBLEU for the four selectable candidates and the submitted MBR
selector. Bold marks the strongest complete candidate in each dialect; $\Delta$ is MBR minus that
candidate before rounding. The final row retrospectively chooses one complete candidate per
dialect using blind-test scores and was not used in system development.}
\label{tab:st1bydialect}
\end{table*}

\begin{figure*}[t]
\centering
\includegraphics[width=0.94\textwidth]{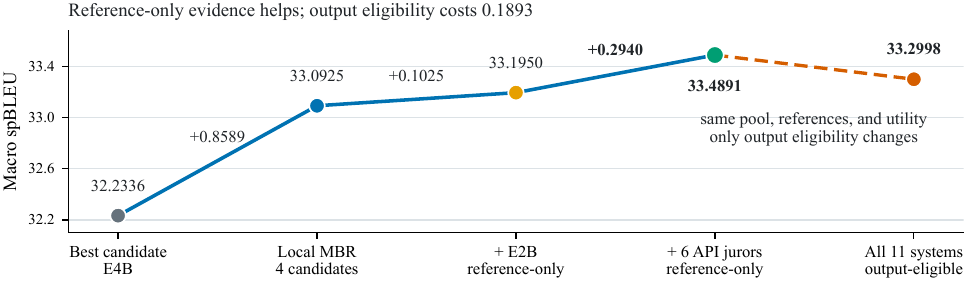}
\caption{Candidate combination, reference evidence, and output eligibility have different effects.
The horizontal path first combines the four candidates and then adds reference-only evidence. The
final branch keeps the complete pool and scoring rule fixed but grants every juror output
eligibility; this post-evaluation control scores 0.1893 below the submitted restricted selector.}
\label{fig:authority}
\end{figure*}

\paragraph{Reference evidence versus output eligibility.}
Figure~\ref{fig:authority} distinguishes candidate combination, added jury evidence, and output eligibility. Combining the four candidates raises macro spBLEU from the strongest standalone score of 32.2336 to 33.0925. Adding reference-only jurors then raises it to 33.1950 with E2B and 33.4891 with the six API jurors. A matched post-evaluation control keeps the full 11-system pool, utility, and deduplication fixed but allows every system to supply the output; it scores 33.2998, forfeiting 0.1893 of the restricted selector's gain. Juror evidence helps under both eligibility regimes, but restricting outputs to the four validated local candidates preserves more of that benefit.

When all 11 systems are eligible, jurors provide 4,893 of 14,459 outputs. Restriction improves 8 of 13 countries, led by OM +1.59, YE +0.72, and LB +0.50; SD is the largest exception, with unrestricted selection ahead by 0.81. The unrestricted control still exceeds both four-candidate MBR and the E2B-only jury, confirming that the result concerns the degree of benefit rather than whether juror evidence helps at all.

\paragraph{Post-evaluation candidate replacement.}
We replace the original NileChat candidate with the full-data soup from Appendix~\ref{app:st1constrained}, keeping the other three candidates, seven reference-only jurors, and selection procedure unchanged. Macro spBLEU rises from 33.4891 to 33.5246 and chrF++ from 48.1320 to 48.1811. The replacement is selected on 14.8\% of turns, compared with 12.6\% for the original candidate, and spBLEU improves in eight of thirteen dialects. Because selectable candidates also supply reference hypotheses, this comparison measures the combined effect of replacing an output option and its reference evidence.

\section{Subtask 2: Retrieval Details and Integrity}
\label{app:st2}

\subsection{Retrieval configuration}
\label{app:st2retrieval}

For a query $q$ from dialect $A$ to dialect $B$, the candidate bank contains the unique released
blind-test sources written in $B$. We encode the bank with multilingual E5-Large using mean pooling
and rank each candidate $c$ by the relative margin
\begin{equation}
  m(q,c)=\frac{2\cos(q,c)}{\bar{s}_4(q,B)+\bar{s}_4(c,A)},
  \label{eq:st2margin}
\end{equation}
where $\bar{s}_4(q,B)$ is the mean similarity between $q$ and its four nearest sources in $B$, and
$\bar{s}_4(c,A)$ is defined symmetrically over sources in $A$. Before ranking, we discard $c$ if
$q$ occurs among its 32 nearest dialect-$A$ sources. This reciprocal filter reduces the chance of
retrieving a cross-dialect counterpart using only source-side similarities. The top $k$ survivors
are supplied as unpaired examples of target-dialect usage, without cross-dialect alignment. This
resembles monolingual target-side retrieval for MT \citep{bouthors-etal-2025-improving}, but the bank
here consists only of released source messages.

Generation uses greedy decoding from Gemma-4-31B-IT. The submitted prompt asks the model to weigh three
renderings internally: one preserving the source closely, one favoring natural target-dialect
expression, and one aligned with the retrieved usage examples. This is a single-call latent variant
inspired by reasoning prompts for MT, not explicit tree search \citep{nguyen-xu-2025-reasoning}.
It returns only one synthesized translation. An exact-copy guard regenerates an output when its
normalized text copies an active example; persistent copies use the Gemma-4-31B-IT zero-shot output
produced without retrieval.

\subsection{Official system ladder}
\label{app:st2ladder}

Table~\ref{tab:st2ladder} reports the complete official blind-test ladder. Scores increase at every
submitted retrieval depth. The $k{=}8$ system sends copied outputs directly to the no-retrieval
fallback, whereas the $k{=}16$ and $k{=}32$ systems first remove the copied example, promote the
next-ranked candidate, and regenerate. Consequently, $k{=}16$ to $k{=}32$ is the controlled depth
comparison. With $k{=}32$ retrieval and repair fixed, guided generation (latent deliberation) adds 0.45 spBLEU (27.9496
to 28.4011).

\par\medskip\noindent
\begin{minipage}{\columnwidth}
\centering
\small
\setlength{\tabcolsep}{3pt}
\begin{tabular}{@{}>{\raggedright\arraybackslash}p{0.61\columnwidth}rr@{}}
\toprule
System & spBLEU & chrF++ \\
\midrule
Copy source unchanged & 12.81 & 27.31 \\
Gemma-4-12B-IT zero-shot & 23.7949 & 37.8899 \\
Gemma-4-31B-IT zero-shot & 24.6898 & 38.8862 \\
Retrieval, $k{=}8$, direct fallback & 26.8595 & 40.7924 \\
Retrieval, $k{=}16$, regenerate & 27.4687 & 41.2666 \\
Retrieval, $k{=}32$, regenerate & 27.9496 & 41.7926 \\
\textbf{$k{=}32$ + guided generation} & \textbf{28.4011} & \textbf{42.1509} \\
\bottomrule
\end{tabular}
\captionof{table}{Official Subtask 2 blind-test ladder. Scores use FLORES-200-tokenized spBLEU and
chrF++; the submitted system is bold.}
\label{tab:st2ladder}
\end{minipage}
\par

\subsection{Development provenance and retrospective audit}
\label{app:st2audit}

Subtask 2 provided no dedicated training or development split. We used public ArBanking77 data
\citep{jarrar-etal-2023-arbanking77} only for exploratory retriever selection. After finding that
some rows align with the benchmark, we excluded measurements on those rows from independent
performance claims. The final retrieval bank and every headline comparison instead use the
released blind-test sources and official blind-test scores, respectively; generation does not
query ArBanking77, hidden references, intent labels, or identifiers linking corresponding rows.

Table~\ref{tab:st2audit} separates the exact-copy handling performed during generation from the
later counterpart audit. In the submitted run, regeneration removed the copied example and
promoted the next-ranked candidate; persistent copies used the no-retrieval Gemma-4-31B-IT output.

\par\medskip\noindent
\begin{minipage}{\columnwidth}
\centering
\small
\setlength{\tabcolsep}{4pt}
\begin{tabular}{@{}>{\raggedright\arraybackslash}p{0.57\columnwidth}
                    >{\raggedleft\arraybackslash}p{0.36\columnwidth}@{}}
\toprule
Event & Incidence \\
\midrule
Outputs initially copying an active example & 30/3,574 (0.84\%) \\
Non-copy after regeneration & 26/30 flagged (86.7\%) \\
No-retrieval fallback & 4/3,574 (0.11\%) \\
Known counterpart collisions, $k{=}8$ prompts & 0/3,574 \\
Known counterpart collisions, final $k{=}32$ prompts & 2/3,574 (0.06\%) \\
Outputs reusing counterpart wording & 1/3,574 (0.03\%) \\
\bottomrule
\end{tabular}
\captionof{table}{Copy handling and retrospective counterpart audit. Rates use all 3,574 outputs
or prompts unless marked as conditional on a flagged output. Counterpart counts include only exact
matches recoverable from public data.}
\label{tab:st2audit}
\end{minipage}
\par\medskip

Only prompts with recoverable public counterparts could be audited, so two is a lower bound on
exact counterpart exposure and the audit cannot establish semantic independence. The reciprocal
filter and output guard reduce counterpart exposure and verbatim copying, but they do not detect
all semantic overlap.

\section{Subtask 3: Controller Analysis and Error Diagnosis}
\label{app:st3}

\subsection{Task calibration and scored systems}

The released labels are strongly imbalanced: sociolinguistic errors account for 57.9\% of the
2,240 training and development spans, followed by semantic errors at 24.4\%. On the 138-row
development set, a trivial no-model baseline that marks the entire Arabic translation as a
sociolinguistic error scores 0.3687 Overall, above the organizers' NileChat-3B baseline at 0.2612.
Its exact-span F1 is only 0.0474, however. The result exposes an important property of the task:
broad character coverage and the majority category can produce a deceptively strong score, while
useful systems must also localize errors precisely.

Table~\ref{tab:st3all} lists our officially scored blind-test systems. The submitted controller
starts from GPT-5.5 semantic proposals, applies MARBERTv2 boundary control, admits disjoint Gemini
proposals only with independent encoder support, rescues empty rows through a fixed fallback chain,
and merges adjacent same-category fragments. Figure~\ref{fig:st3panels} summarizes the resulting
score progression and the pre-test error diagnosis.

The four-system character-probability combination instead pools GPT-5.5, MARBERTv2, Gemini 3.6 Flash, and the Gemma-4-12B-IT SFT extractor. It retains characters covered by at least two systems, joins contiguous retained characters into spans, assigns each span a category by voting among overlapping predictions, and merges adjacent same-category fragments. This alternative scores 0.4825, below the submitted controller's 0.4982.

\subsection{Validation transfer}
\label{app:st3validation}

Pre-test decisions used stratified five-fold cross-validation over all 1,263 released training and development rows, with tagger retraining and fold-aware demonstration selection in every fold. The fully local Gemma-4-31B-IT/MARBERTv2 span-union hybrid averaged 0.4549 Overall across folds (standard deviation 0.0163); pooled out-of-fold scoring gave 0.4573.

Transfer depended strongly on how a change was validated. A character-level union measured against the exact span-union baseline it replaced improved cross-validation by 0.0084 (95\% CI [+0.0049, +0.0123]) and retained a +0.0037 blind-test gain. On the development configuration used to select the submitted extension (cached GPT-5.5 proposals and selected MARBERTv2 tagger), extended widening and agreement-gated additions raised Overall from 0.4374 to 0.4567 (+0.0193). The configuration differs from Table~\ref{tab:st3mpp}'s metric-comparison diagnostic. The submitted extension, which also included empty-row rescue, improved blind-test Overall by 0.0181. Fragment merging shrank from +0.0069 in cross-validation to +0.0015 on the blind test. By contrast, changes selected on the 138-row development set or direction-level slices of roughly 27 rows often reversed. These results favor matched comparisons on the largest available labeled split over selecting small global gains from a single development partition.

\begin{table*}[t]
\centering
\small
\resizebox{0.92\textwidth}{!}{\begin{tabular}{l c r r r}
\toprule
System & API models & Overall & Overlap F1 & Class F1 \\
\midrule
\textbf{Submitted controller} & 2 & \textbf{0.4982} & 0.5426 & 0.4538 \\
Controller without fragment merge & 2 & 0.4967 & 0.5439 & 0.4494 \\
Four-system character-probability combination & 2 & 0.4825 & 0.5309 & 0.4342 \\
GPT-5.5 + boundary control & 1 & 0.4786 & 0.5175 & 0.4396 \\
Direction-routed controller & 1 & 0.4772 & 0.5200 & 0.4343 \\
GPT-5.5 proposal & 1 & 0.4691 & 0.5137 & 0.4244 \\
Gemini 3.6 Flash + boundary control & 1 & 0.4456 & 0.4890 & 0.4022 \\
\underline{Local Gemma-4-31B-IT + MARBERTv2, character union} & 0 & 0.4446 & 0.5045 & 0.3847 \\
Local Gemma-4-31B-IT + MARBERTv2, span union & 0 & 0.4409 & 0.5048 & 0.3770 \\
Gemma-4-12B-IT SFT + MARBERTv2 & 0 & 0.4357 & 0.5062 & 0.3651 \\
MARBERTv2, three-seed ensemble & 0 & 0.3817 & 0.4535 & 0.3099 \\
\bottomrule
\end{tabular}
}
\caption{Official blind-test scores for our Subtask 3 systems. The submitted controller is bold;
the underlined row is the strongest system without API models.}
\label{tab:st3all}
\end{table*}

\subsection{Why span discovery remains the bottleneck}

Cross-model agreement separates gold-matched from spurious proposed spans at only 0.70 to 0.74 area
under the ROC curve (AUC)
on the offline splits, which is too weak for standalone filtering. We therefore use agreement only
to gate disjoint additions. Two post-evaluation controls confirm that broader consensus does not
recover the missing evidence: extending submitted spans to character regions supported by at least
two other systems scores 0.4951, and span-set MBR over all eleven scored systems reaches 0.4897,
both below the submitted 0.4982. Consensus can adjudicate among proposed spans, but it cannot recover
an error region that no component proposed.

\paragraph{Metric sensitivity.}
The deployed scorer averages character-overlap F1 and pooled class F1. We also re-score the
controller stages with partial-overlap, partial-credit matching and micro-averaging (MPP-micro), a
recommended robust span-evaluation regime \citep{perrella-etal-2026-span}. Table~\ref{tab:st3mpp}
shows that agreement-gated additions and fragment merging retain positive gains under MPP-micro.
Boundary widening is more metric-dependent: its cross-validation gain falls from +0.0090 under the
deployed scorer to +0.0019 under MPP-micro because added characters improve mask coverage but reduce
pair precision. The diagnosis itself is stable: unmatched and missed spans remain substantially
more common than wrong categories among matched spans.

\subsection{Error analysis}

\begin{figure*}[p]
\centering
\includegraphics[width=0.94\textwidth]{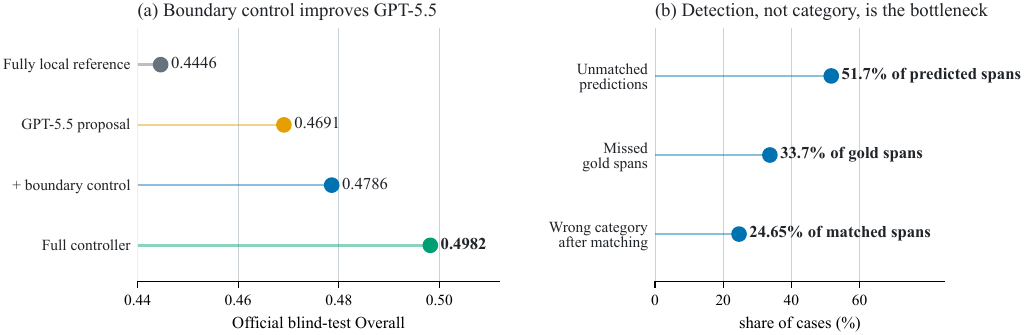}
\caption{Subtask 3 controller analysis. (a) Official blind-test Overall for the fully local character-union hybrid and three proposer-controller stages. (b) Five-fold out-of-fold error rates for the pre-test direction-routed system: the local Gemma-4-31B-IT/MARBERTv2 hybrid handles Mauritanian inputs, and GPT-5.5 with boundary widening handles the other directions. Each rate uses the printed denominator. Unmatched predictions and missed gold spans remain more common than category errors among already matched spans.}
\label{fig:st3panels}
\end{figure*}

Table~\ref{tab:st3confusion} analyzes GPT-5.5 proposals with MARBERTv2 boundary widening on the 1,263 cross-validation rows, rather than the direction-routed system in Figure~\ref{fig:st3panels}(b). Of 2,240 gold spans, 772 receive no overlapping prediction; 1,773 of 3,241 predictions match no gold span. Once a span is matched, its category is correct in 1,100 of 1,468 cases (74.9\%). The remaining confusions concentrate in the two head categories: 247 of 368 matched errors are assigned semantics or sociolinguistics. This again identifies span discovery as the larger bottleneck.

\par\medskip\noindent\begin{minipage}{\columnwidth}
\paragraph{Missed error.}
[ENG\_MAU, id 60306\_9] Source: \emph{``If not, they will see what they can do.''}
MT:~\raisebox{-7pt}{\includegraphics{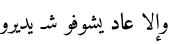}}
(English gloss: ``Otherwise, they will see what they can do.'')
(gold: the bold word, sociolinguistics; predicted: no span). The error is a single discourse particle in an
otherwise dialect-looking sentence, so no controller stage can recover it after every proposer
misses it.
\end{minipage}

\paragraph{Most frequent category confusion.}
[ENG\_EGY, id 1199\_1] Source: \emph{``Did you hear? You heard his frankness; he is straight
and honest.''} MT:~\raisebox{-7pt}{\includegraphics{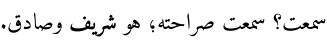}}
(English gloss: ``Did you hear? You heard his frankness; he is honorable and
honest.'') (gold: sociolinguistics; predicted: semantics). The proposer treats
the bold word (``honorable'' for ``straight'') as a lexical-choice error, whereas the gold
annotation treats it as a register or variety error. This sociolinguistics-to-semantics cell
contains 90 of the 368 matched category errors.

\paragraph{Boundary widening can reduce precision.}
[ENG\_EGY, id 6181\_1] Source: \emph{``Or at least remove the worry from his heart.''}
MT:~\raisebox{-7pt}{\includegraphics{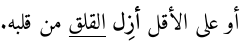}}
(English gloss: ``Or at least remove the worry from his heart.'')
(gold and base prediction: the bold word, sociolinguistics). MARBERTv2 widens the exact base span to
include the underlined word, adding six false-positive characters and reducing row-level character-mask
F1 from 1.0 to 0.571. This is the example-level counterpart of widening's metric sensitivity in
Table~\ref{tab:st3mpp}.

\begin{table*}[p]
\centering
\small
\renewcommand{\arraystretch}{1.10}
\begin{tabular}{llrrrr}
\toprule
Split & Stage & Deployed Overall & $\Delta$ & MPP-micro & $\Delta$ \\
\midrule
CV & GPT-5.5 proposal & 0.4565 & & 0.4659 & \\
CV & + boundary widening & 0.4655 & +0.0090 & 0.4678 & +0.0019 \\
CV & + fragment merge & 0.4724 & +0.0069 & 0.4773 & +0.0095 \\
CV & local reference (span union) & 0.4573 & & 0.4610 & \\
\midrule
dev & GPT-5.5 proposal & 0.4236 & & 0.4315 & \\
dev & + boundary widening & 0.4296 & +0.0059 & 0.4301 & $-$0.0014 \\
dev & + agreement-gated additions & 0.4459 & +0.0163 & 0.4500 & +0.0199 \\
dev & + fragment merge & 0.4494 & +0.0035 & 0.4514 & +0.0014 \\
\bottomrule
\end{tabular}
\caption{Controller stages under the deployed scorer and MPP-micro on released rows with gold (CV: 1,263 rows; dev: 138 rows). Deltas compare consecutive stages within each block. The local reference is the Gemma-4-31B-IT/MARBERTv2 span-union hybrid scored on pooled out-of-fold predictions, not part of the preceding CV chain.}
\label{tab:st3mpp}

\vspace{1.5em}

\begin{tabular}{@{}lrrrrrrrr@{}}
\toprule
& \multicolumn{6}{c}{Predicted category} & & \\
\cmidrule(lr){2-7}
Gold & Gra & Mor & Ort & Pra & Sem & Soc & Missed & Support \\
\midrule
Graphetics       & 1 & 0  & 0  & 0  & 0   & 0    & 1   & 2 \\
Morphosyntax     & 0 & 57 & 3  & 0  & 31  & 13   & 82  & 186 \\
Orthography      & 0 & 16 & 32 & 0  & 21  & 5    & 43  & 117 \\
Pragmatics       & 0 & 1  & 1  & 4  & 30  & 14   & 41  & 91 \\
Semantics        & 0 & 28 & 19 & 17 & 226 & 43   & 213 & 546 \\
Sociolinguistics & 0 & 17 & 10 & 9  & 90  & 780  & 392 & 1298 \\
\midrule
Spurious pred.   & 12 & 137 & 96 & 23 & 370 & 1135 & & 1773 \\
\bottomrule
\end{tabular}
\caption{Category confusion for GPT-5.5 proposals with MARBERTv2 boundary widening on 1,263 cross-validation rows, using greedy one-to-one matching for spans with nonzero character overlap. Gra, Mor, Ort, Pra, Sem, and Soc abbreviate the six LQM categories.}
\label{tab:st3confusion}
\end{table*}

\twocolumn[{\begin{minipage}{\textwidth}
\section{Cross-Task Comparison Record}
\label{app:comparisonrecord}

\begingroup
\centering\small
\setlength{\tabcolsep}{4pt}
\begin{tabular}{@{}l>{\raggedright\arraybackslash}p{0.58\textwidth}rrc@{}}
\toprule
ID & Configuration & Score & $\Delta$ (vs. row) & Phase \\
\midrule
\multicolumn{5}{@{}l}{\textit{Unconstrained ST1: fixed local candidates}} \\
U0 & Strongest standalone candidate (E4B) & 32.2336 & n/a & S \\
U1 & Four-candidate MBR without additional jurors & 33.0925 & +0.8589 (U0) & P \\
U2 & + E2B reference-only juror & 33.1950 & +0.1025 (U1) & S \\
U3 & + Six API reference-only jurors (submitted) & \textbf{33.4891} & +0.2940 (U2) & S \\
U4 & All eleven systems made output-eligible & 33.2998 & -0.1893 (U3) & P \\
\addlinespace[4pt]
\multicolumn{5}{@{}l}{\textit{Unconstrained ST1: candidate replacement}} \\
U5 & Submitted MBR with the original NileChat candidate replaced by the full-data three-seed soup & 33.5246 & +0.0355 (U3) & P \\
\addlinespace[4pt]
\multicolumn{5}{@{}l}{\textit{Constrained ST1: submitted routing}} \\
C0 & Three-seed NileChat soup for all countries & 29.5969 & n/a & S \\
C1 & Route LY/SD to train + public-test adapter (submitted) & \textbf{30.4233} & +0.8264 (C0) & S \\
\addlinespace[4pt]
\multicolumn{5}{@{}l}{\textit{Constrained ST1: supervision and weight averaging}} \\
C2 & Train + development & 29.2323 & n/a & P \\
C3 & + Duplicated rows at LY/SD budget & 29.0971 & -0.1352 (C2) & P \\
C4 & + Authentic public LY/SD rows & 30.1179 & +0.8856 (C2) & P \\
C5 & + Duplicated rows at full-public-test budget & 28.8483 & -0.3840 (C2) & P \\
C6 & + Authentic complete public test & 30.4733 & +1.2410 (C2) & P \\
C7 & Equal-weight soup of three full-data models & 30.9726 & +0.4993 (C6) & P \\
\addlinespace[4pt]
\multicolumn{5}{@{}l}{\textit{ST2: Gemma-4-31B-IT without fine-tuning}} \\
T0 & Zero-shot without retrieval & 24.6898 & n/a & S \\
T1 & Target-dialect retrieval, $k{=}32$, with copy guard & 27.9496 & +3.2597 (T0) & S \\
T2 & + Guided generation (submitted) & \textbf{28.4011} & +0.4516 (T1) & S \\
\addlinespace[4pt]
\multicolumn{5}{@{}l}{\textit{ST3: semantic proposals and span control}} \\
E0 & Raw GPT-5.5 proposals & 0.4691 & n/a & S \\
E1 & + MARBERTv2 boundary widening & 0.4786 & +0.0095 (E0) & S \\
E2 & + Gemini widening / gated additions / empty-row rescue & 0.4967 & +0.0181 (E1) & S \\
E3 & + Same-category fragment merging (submitted) & \textbf{0.4982} & +0.0015 (E2) & S \\
\bottomrule
\end{tabular}
\captionof{table}{Cross-task blind-test comparisons. S: official evaluation-window submission; P: unofficial post-evaluation experiment on the same blind test. Scores are macro spBLEU for ST1, spBLEU for ST2, and Overall for ST3. Each $\Delta$ names its reference row. C3 to C6 independently modify C2; all five use seed-42 adapters. C7 averages three full-data models, distinct from C0. ST2 retrieval includes copy handling; E2 combines controller extensions. Effects are not independently additive. Deltas use unrounded scores except for ST3's four-decimal records. Submitted final scores are bold.}
\label{tab:comparisonrecord}
\par\endgroup

\end{minipage}}]

\end{document}